%% file: PaperForReview.tex
\documentclass[10pt,twocolumn,letterpaper]{article}

\usepackage[pagenumbers]{wacv} 

\usepackage{listings}
\input{src/x0.Preamble.tex}
\input{src/x1.math_commands.tex}
\input{src/x2.symbols.tex}

\usepackage[pagebackref,breaklinks,colorlinks]{hyperref}

\usepackage[capitalize]{cleveref}
\crefname{section}{Sec.}{Secs.}
\Crefname{section}{Section}{Sections}
\Crefname{table}{Table}{Tables}
\crefname{table}{Tab.}{Tabs.}

\def\wacvPaperID{1273} 
\def\confName{WACV}
\def\confYear{2024}

\begin{document}

\input{src/00.TitleAuthors}

\maketitle

\begin{abstract}
   \input{src/00.Abstract.tex}
\end{abstract}

\setlength{\abovedisplayskip}{3pt}
\setlength{\belowdisplayskip}{3pt}

\addtocontents{toc}{\protect\setcounter{tocdepth}{0}}
\input{src/01.Introduction.tex}
\input{src/02.Background.tex}
\input{src/03.Methods.tex}

\input{src/04.Experiments.tex}

\input{src/05.Discussion.tex}

{\small
\bibliographystyle{ieee_fullname}
\bibliography{PaperForReview}
}

\addtocontents{toc}{\protect\setcounter{tocdepth}{1}}
\appendix
\input{src/suppmat}

\end{document}

%% file: src/x0.Preamble.tex
\usepackage{times}
\usepackage{epsfig}
\usepackage{graphicx}
\usepackage{amsmath}
\usepackage{amssymb}
\usepackage[dvipsnames]{xcolor}

\usepackage[hyphens]{url}  
\usepackage{booktabs}
\usepackage{multirow}
\usepackage{float}
\usepackage{subcaption}
\usepackage{enumitem}
\usepackage{stmaryrd}
\usepackage{bbm}
\usepackage{pifont}
\usepackage{makecell}
\usepackage[
  ruled,
  vlined,
  linesnumbered,
]{algorithm2e}

\newcommand{\ProtoNetStyle}[1]{{\small\textsc{#1}}}

\newcommand{\Ours}{\ProtoNetStyle{PixPNet}}

\newcommand{\ProtoPNet}{\ProtoNetStyle{ProtoPNet}}
\newcommand{\ProtoPool}{\ProtoNetStyle{ProtoPool}}
\newcommand{\ProtoPShare}{\ProtoNetStyle{ProtoPShare}}
\newcommand{\TesNet}{\ProtoNetStyle{TesNet}}

\newcommand{\ProSeNet}{\ProtoNetStyle{ProSeNet}}
\newcommand{\ViTNeT}{\ProtoNetStyle{ViT-NeT}}
\newcommand{\ProtoTree}{\ProtoNetStyle{ProtoTree}}
\newcommand{\ProtoPMedEEG}{\ProtoNetStyle{ProtoPMed-EEG}}
\newcommand{\ProtGNN}{\ProtoNetStyle{ProtGNN}}
\newcommand{\PxGNN}{\ProtoNetStyle{PxGNN}}
\newcommand{\DeformableProtoPNet}{\ProtoNetStyle{Deformable ProtoPNet}}
\newcommand{\ProtoToProto}{\ProtoNetStyle{Proto2Proto}}
\newcommand{\ProtoPFormer}{\ProtoNetStyle{ProtoPFormer}}
\newcommand{\ProtoSeg}{\ProtoNetStyle{ProtoSeg}}
\newcommand{\STProtoPNet}{\ProtoNetStyle{ST-ProtoPNet}}

\newcommand{\GCNTesNetProtoPNet}{\ProtoNetStyle{GCN-\{TesNet,ProtoPNet\}}}
\newcommand{\QuasiProtoPNet}{\ProtoNetStyle{Quasi-ProtoPNet}}
\newcommand{\NPProtoPNet}{\ProtoNetStyle{NP-ProtoPNet}}
\newcommand{\PsProtoPNet}{\ProtoNetStyle{Ps-ProtoPNet}}
\newcommand{\GenProtoPNet}{\ProtoNetStyle{Gen-ProtoPNet}}
\newcommand{\ProtoLNet}{\ProtoNetStyle{ProtoLNet}}
\newcommand{\ProtoVAE}{\ProtoNetStyle{ProtoVAE}}
\newcommand{\DPNet}{\ProtoNetStyle{DPNet}}
\newcommand{\XProtoNet}{\ProtoNetStyle{XProtoNet}}
\newcommand{\SemiProtoPNet}{\ProtoNetStyle{Semi-ProtoPNet}}
\newcommand{\SDFASAProtoPNet}{\ProtoNetStyle{SDFA-SA-ProtoPNet}}
\newcommand{\ProtoPDebug}{\ProtoNetStyle{ProtoPDebug}}
\newcommand{\PRP}{\ProtoNetStyle{PRP}}

\newcommand{\ResNet}[1][]{\ProtoNetStyle{ResNet#1}}
\newcommand{\VGG}[1][]{\ProtoNetStyle{VGG#1}}
\newcommand{\DenseNet}[1][]{\ProtoNetStyle{DenseNet#1}}

\newcommand{\TabProtoNetStyle}[1]{{\footnotesize\textsc{#1}}}

\newcommand{\TabOurs}{\TabProtoNetStyle{PixPNet}}

\newcommand{\TabProtoPNet}{\TabProtoNetStyle{ProtoPNet}}
\newcommand{\TabProtoPool}{\TabProtoNetStyle{ProtoPool}}
\newcommand{\TabProtoPShare}{\TabProtoNetStyle{ProtoPShare}}
\newcommand{\TabTesNet}{\TabProtoNetStyle{TesNet}}

\newcommand{\TabViTNeT}{\TabProtoNetStyle{ViT-NeT}}
\newcommand{\TabProtoTree}{\TabProtoNetStyle{ProtoTree}}

\newcommand{\TabSupportProtoPNet}{\TabProtoNetStyle{Support ProtoPNet}}
\newcommand{\TabProtoToProto}{\TabProtoNetStyle{Proto2Proto}}
\newcommand{\TabProtoPFormer}{\TabProtoNetStyle{ProtoPFormer}}

\newcommand{\TabSTProtoPNet}{\TabProtoNetStyle{ST-ProtoPNet}}

\newcommand{\TabPRP}{\TabProtoNetStyle{PRP}}

\newcommand{\Protonet}{ProtoPartNN}
\newcommand{\Protonets}{\Protonet{}s}

\newcommand{\RFAlg}{{\small\texttt{FunctionalRF}}}

\makeatletter
\renewcommand\paragraph{\@startsection{paragraph}{4}{\z@}%
  {0.5ex \@plus1ex \@minus.2ex}%
  {-1em}%
  {\normalfont\normalsize\bfseries}}
\makeatother

\newcommand{\fakeparagraph}[1]{\vspace{.2ex}\noindent\textit{#1}~~}

\newcommand{\good}[1]{\textcolor{ForestGreen}{#1}}
\newcommand{\bad}[1]{\textcolor{BrickRed}{#1}}

\newcommand{\cmark}{\good{\ding{51}}}
\newcommand{\xmark}{\bad{\ding{55}}}

\newcommand{\cmarkInv}{\bad{\ding{51}}}
\newcommand{\xmarkInv}{\good{\ding{55}}}

%% file: src/x1.math_commands.tex
\usepackage{amsmath,amsfonts,bm,mathtools}

\def\eqref#1{Eq.~(\ref{#1})}

\def\1{\bm{1}}

\def\vp{{\bm{p}}}

\def\vs{{\bm{s}}}

\def\vx{{\bm{x}}}
\def\vy{{\bm{y}}}
\def\vz{{\bm{z}}}

\def\mM{{\bm{M}}}

\def\mP{{\bm{P}}}

\def\mS{{\bm{S}}}

\def\mW{{\bm{W}}}

\def\mY{{\bm{Y}}}
\def\mZ{{\bm{Z}}}

\DeclareMathAlphabet{\mathsfit}{\encodingdefault}{\sfdefault}{m}{sl}
\SetMathAlphabet{\mathsfit}{bold}{\encodingdefault}{\sfdefault}{bx}{n}
\newcommand{\tens}[1]{\bm{\mathsfit{#1}}}

\def\tD{{\tens{D}}}

\def\tX{{\tens{X}}}

\def\gG{{\mathcal{G}}}

\def\sR{{\mathbb{R}}}

\def\emS{{S}}

\newcommand{\Ls}{\mathcal{L}}

\DeclareMathOperator*{\argmin}{arg\,min}

\DeclarePairedDelimiterX{\norm}[1]{\lVert}{\rVert}{#1}
\newcommand{\pnorm}[2]{\norm{#2}_#1}

%% file: src/x2.symbols.tex
\def\prototypes{{\mP}}
\def\prototype{{\vp}}
\def\latentz{{\vz}}
\def\latentZ{{\mZ}}
\def\similarities{{\vs}}
\def\similarity{{s}}
\def\sample{{\vx}}
\def\dataset{{\tD}}
\def\samples{{\tX}}
\def\ylabels{{\mY}}
\def\logits{{\hat{\vy}}}
\def\prediction{{\hat{y}}}
\def\ylabel{{y}}
\def\distancef{{\varphi}}
\def\lossTotal{{\Ls_{\text{total}}}}
\def\lossXent{{\Ls_{\text{xent}}}}
\def\lossSep{{\Ls_{\text{sep}}}}
\def\lossSepCoef{{\lambda_{\text{sep}}}}
\def\lossCls{{\Ls_{\text{cls}}}}
\def\lossClsCoef{{\lambda_{\text{cls}}}}
\def\lossReadout{{\Ls_{h}}}
\def\lossReadoutCoef{{\lambda_{h}}}
\def\featuref{{f}}
\def\backbone{{f_{\text{core}}}}
\def\addonf{{f_{\text{add}}}}
\def\protof{{g}}
\def\simf{{v}}
\def\simmap{{\mS}}
\def\simmapf{{\pi}}
\def\readoutf{{h}}
\def\patches{{\small\texttt{patches}}}
\def\readoutW{{\mW_h}}
\def\readoutw{{w_h}}

%% file: src/00.TitleAuthors.tex
\title{Pixel-Grounded Prototypical Part Networks}

\author{Zachariah Carmichael\textsuperscript{1,2} \quad Suhas Lohit\textsuperscript{2} \quad Anoop Cherian\textsuperscript{2} \quad Michael J. Jones\textsuperscript{2} \quad Walter J. Scheirer\textsuperscript{1}\\
\textsuperscript{1}University of Notre Dame\\
\textsuperscript{2}Mitsubishi Electric Research Laboratories\\
{\tt\small zcarmich@nd.edu, \{slohit,anoop.cherian,mjones\}@merl.com, walter.scheirer@nd.edu}
}

%% file: src/00.Abstract.tex
Prototypical part neural networks (\Protonets{}), namely \ProtoPNet{} and its derivatives, are an intrinsically interpretable approach to machine learning. Their prototype learning scheme enables intuitive explanations of the form, \emph{this} (prototype) looks like \emph{that} (testing image patch). But, does \emph{this} actually look like \emph{that}? In this work, we delve into why object part localization and associated heat maps in past work are misleading. Rather than localizing to object parts, existing \Protonets{} localize to the entire image, contrary to generated explanatory visualizations. We argue that detraction from these underlying issues is due to the alluring nature of visualizations and an over-reliance on intuition. To alleviate these issues, we devise new receptive field-based architectural constraints for meaningful localization and a principled pixel space mapping for \Protonets{}. To improve interpretability, we propose additional architectural improvements, including a simplified classification head. We also make additional corrections to \ProtoPNet{} and its derivatives, such as the use of a validation set, rather than a test set, to evaluate generalization during training. Our approach, \Ours{} (Pixel-grounded Prototypical part Network), is the \textbf{only} \Protonet{} that truly learns and localizes to prototypical object parts. We demonstrate that \Ours{} achieves quantifiably improved interpretability without sacrificing accuracy.

%% file: src/01.Introduction.tex
\section{Introduction}

Prototypical part neural networks (\Protonets{}) are an attempt to remedy the inscrutability and fundamental lack of trustworthiness characteristic of canonical deep neural networks~\cite{protopnet}. By learning prototypes of object parts, \Protonets{} make human-interpretable predictions with justifications of the form: \emph{this} (training image patch) looks like \emph{that} (testing image patch). Since black-box AI systems often obfuscate their deficiencies~\cite{adversarialExamples,hendrycksNaturalAdversarialExamples2019,liptonMythos2018}, \Protonets{} represent a shift in the direction of transparency. With unprecedented interest in AI from decision-makers in high-stakes industries -- \eg{}, medicine, finance, and law~\cite{liptonMythos2018,millerAIMedical2018,Rudin2019,medicalXAIsurvey2020} -- the demand for explainable AI systems is greater than ever. Further motivation for transparency is driven by real-world consequences of deployed black boxes~\cite{oneilWeaponsMathDestruction2016,buolamwiniGenderShadesIntersectional2018,incidentDB} and mounting regulatory ordinance~\cite{EU-GDPR,EU-US-TTC_statement2021,EU-AI-Act,US-Alg-Acct-Act}.

\begin{figure}
    \centering
    \includegraphics[width=\linewidth]{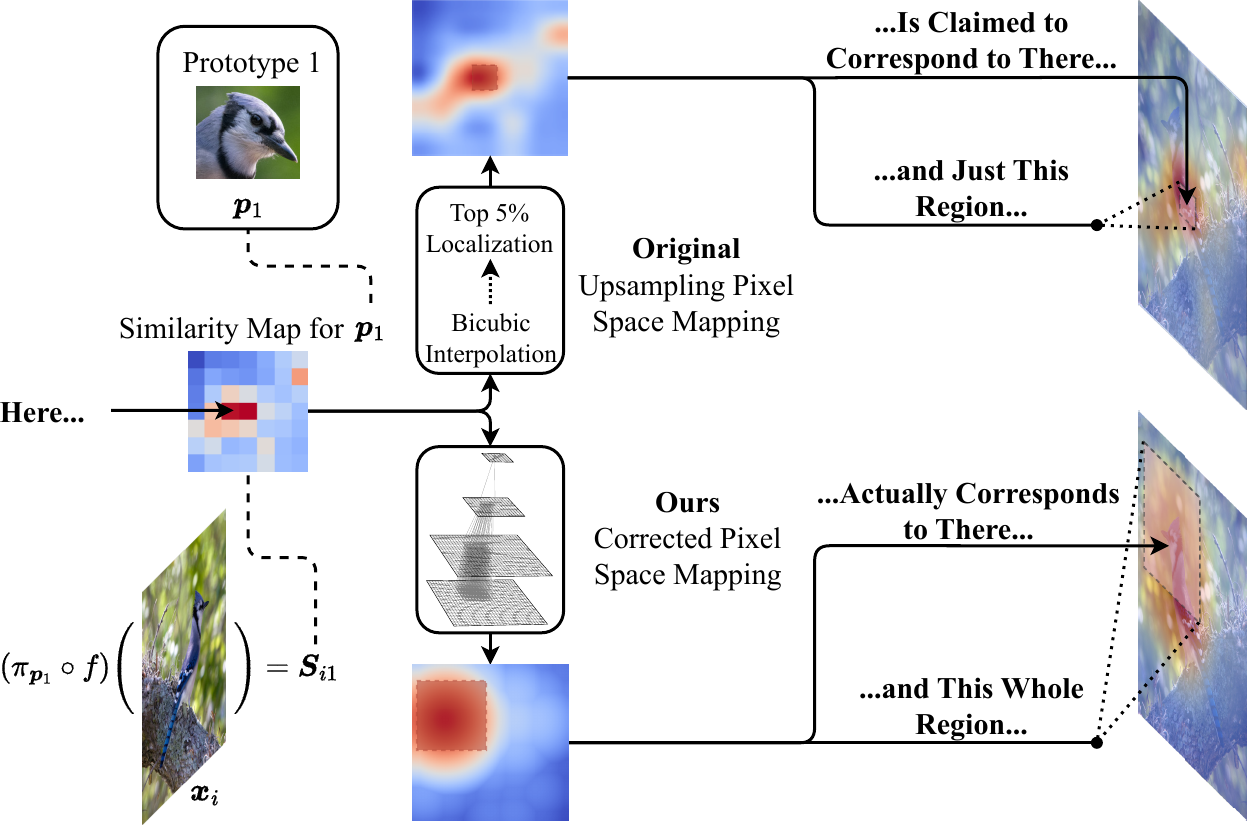}
    \caption{The two primary issues identified with prototype visualization: \textit{here} (this embedded patch) does not correspond to \textit{there} (this image patch), and \textit{this}
    (prototype)
    does not correspond to \textit{just that} (test image patch). In the extreme case, \textit{this} can actually correspond to \textit{the entire image} (\ie{}, when the receptive field is 100\%).
    }
    \label{fig:thisThatHereThere}
\end{figure}

\Protonets{} approach explainability from an intrinsically interpretable lens and offer many benefits over post hoc explanation. Whereas post hoc explainers estimate an explanation, \Protonet{} explanations are part of the actual prediction process -- explanations along the lines of ``\textit{this} looks like \textit{that}'' follow naturally from the symbolic form of the model itself. This implicit explanation is characteristic of models widely considered to be human-comprehensible~\cite{razIMLForm2022}.
Moreover, \Protonets{} enable concept-level debugging, human-in-the-loop learning, and implicit localization~\cite{ExplainingPrototypes,ProSeNet,protopnet}.
Being independent of the explained model, post hoc explainers have been found to
be
unfaithful, inconsistent, and unreliable~\cite{krishna2022disagreement,fooling,10.1145/3442188.3445941,carmichaelposthoceval2021} (see Section~\ref{sec:xai_methods} for expanded discussion).

When misunderstood or used inappropriately, explainable AI (XAI) methods can have unintended consequences~\cite{kaurInterpretingInterpretabilityUnderstanding2020,krishna2022disagreement}.
This harm arises from unverified hypotheses, whether it is that explanations represent phenomena faithful to the predictor or meaningful properties of the predictor.
So, why do we see such
hypotheses
proliferating throughout both academia and industry~\cite{falsifiable2020,kaurInterpretingInterpretabilityUnderstanding2020}?
The problem is very human -- there is often an over-reliance on intuition that may lead to illusory progress or deceptive conclusions. Whether it is dependence on alluring visualization or behavioral extrapolation from cherry-picked examples, XAI methods often are left insufficiently scrutinized and subject to ``researcher degrees of freedom''~\cite{researcherDegsFreedom2011,falsifiable2020}.

Recent evidence indicates that \Protonets{} may suffer from these same issues: 
\ProtoPNet{} and its variants exhibit irrelevant prototypes, a human-machine semantic similarity gap, and exorbitant explanation size~\cite{protoStudy1,protoStudyHIVE,protoAttack}.
Unfortunately, in our study, we confirm that this is the case --
there are several facets of existing \Protonet{} explanations that do \emph{not} result from 
the implicit form of the model: object part localization, pixel space grounding, and heat map visualizations.
Instead, these are founded on unverified assumptions and an over-reliance on intuition, often justified \textit{a posteriori} by attractive visuals.
We demonstrate that, colloquially, \textit{this} does not actually look like \textit{that}, and \textit{here} may not actually correspond to \textit{there} -- see Figure~\ref{fig:thisThatHereThere} for illustration. 
These issues with \Protonets{} are not limited to just \ProtoPNet{}, but to all of its derivatives.

This work aims to elevate the interpretability of \Protonets{} by rectifying these facets. In doing so, all aspects of \Protonet{} explanations are embedded in the symbolic form of the model.
Our contributions are as follows:
\begin{itemize}[noitemsep,nolistsep,leftmargin=1em]
    \item We identify that existing \Protonets{} based on \ProtoPNet{} do not localize faithfully nor actually localize to object parts, but rather the full image in most cases.
    \item We propose a novel pixel space mapping based on the receptive fields of an architecture (we guarantee that \textit{here} corresponds to \textit{there}).
    \item We propose architectural constraints that we efficiently discover through a transfer task to enable true object part localization (\textit{this} looks like \textit{that}).
    \item We devise a novel functional algorithm for the receptive field calculation of any architecture.
    \item On several image classification tasks, our approach, \Ours{}, achieves competitive accuracy with other \Protonets{} \textit{while maintaining a higher degree of interpretability}, as substantiated by functionally grounded XAI metrics, and being the \textit{only \Protonet{} that truly localizes to object parts}.
\end{itemize}

%% file: src/02.Background.tex
\section{Background}\label{sec:xai_methods}

In this section, we give a brief background of explainable AI methods, the \ProtoPNet{} formulation, and an overview of \ProtoPNet{} extensions.

\paragraph{Explainable AI Methods}
Explainable AI (XAI) solutions can be classified as post hoc, intrinsically interpretable, or a hybrid of the two~\cite{Schwalbe2023}. Whereas intrinsically interpretable methods are both the explanator and predictor, post hoc methods act as an explanator for an independent predictor.
Unfortunately, post hoc explainers are known to be inconsistent, unfaithful, and possibly even intractable~\cite{krishna2022disagreement,bordt2022posthoc,DBLP:conf/aaai/BroeckLSS21,pmlr-v108-garreau20a,carmichaelposthoceval2021}.
Furthermore, they are deceivable~\cite{fooling,dimanov2020you,dombrowski2019explanations,adv_xai_site} and have been shown to not affect, or even reduce, end-user task performance \cite{10.1145/3442188.3445941,humer2022comparing}.
While this is the case, post hoc explanations have been shown to possibly increase user trust in AI systems~\cite{carter2022explainable}, improve end-user performance for some explanation types and tasks~\cite{humer2022comparing}, and explain black boxes in trustless auditing schemes~\cite{unfooling2023}.
However, for high-stakes domains, post hoc explanation is frequently argued to be especially inappropriate~\cite{Rudin2019}.

For these numerous reasons, our work concerns intrinsically interpretable machine learning solutions (see \cite{Schwalbe2023} for a methodological overview). In particular, we are interested in \textit{prototypical part neural networks} (\Protonets{})~\cite{protopnet}.
\begin{figure*}
    \centering
    \begin{subfigure}{0.67\linewidth}
        \centering
        \includegraphics[width=\linewidth]{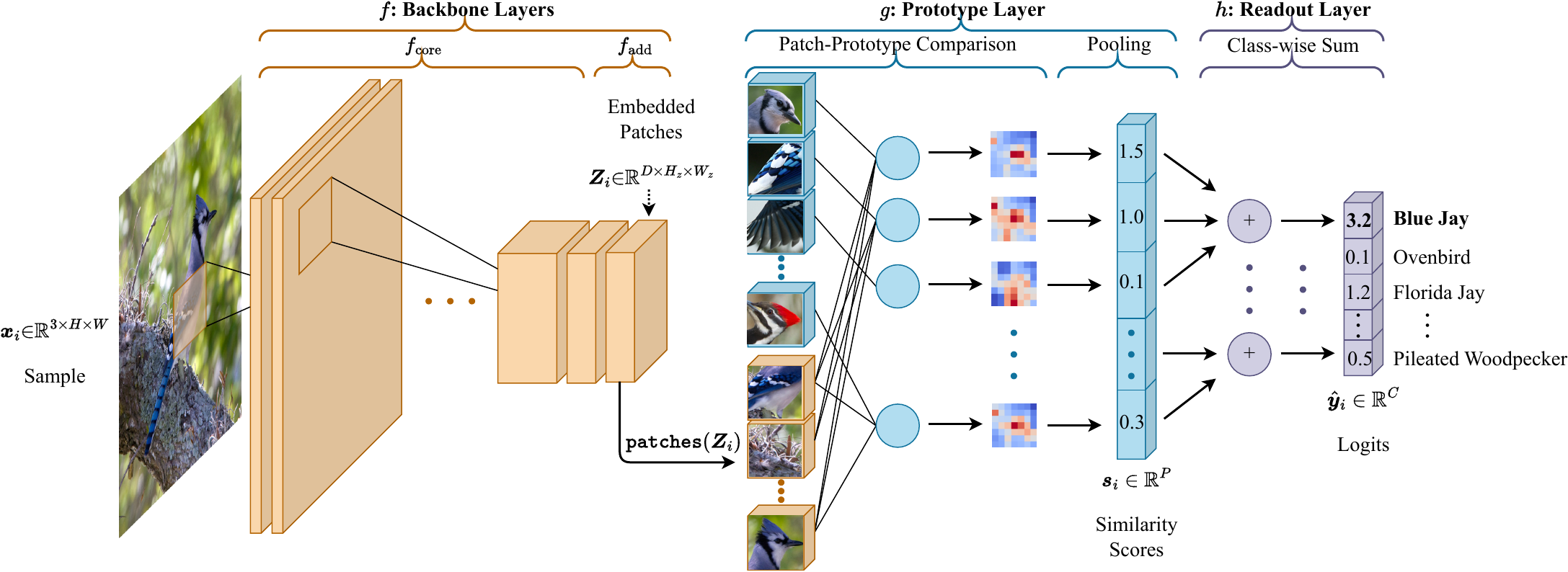}
        \caption{\Ours{} (Pixel-grounded Prototypical part Network) architecture.}
    \end{subfigure}\hfill%
    \begin{subfigure}{0.32\linewidth}
        \centering
        \includegraphics[width=\linewidth]{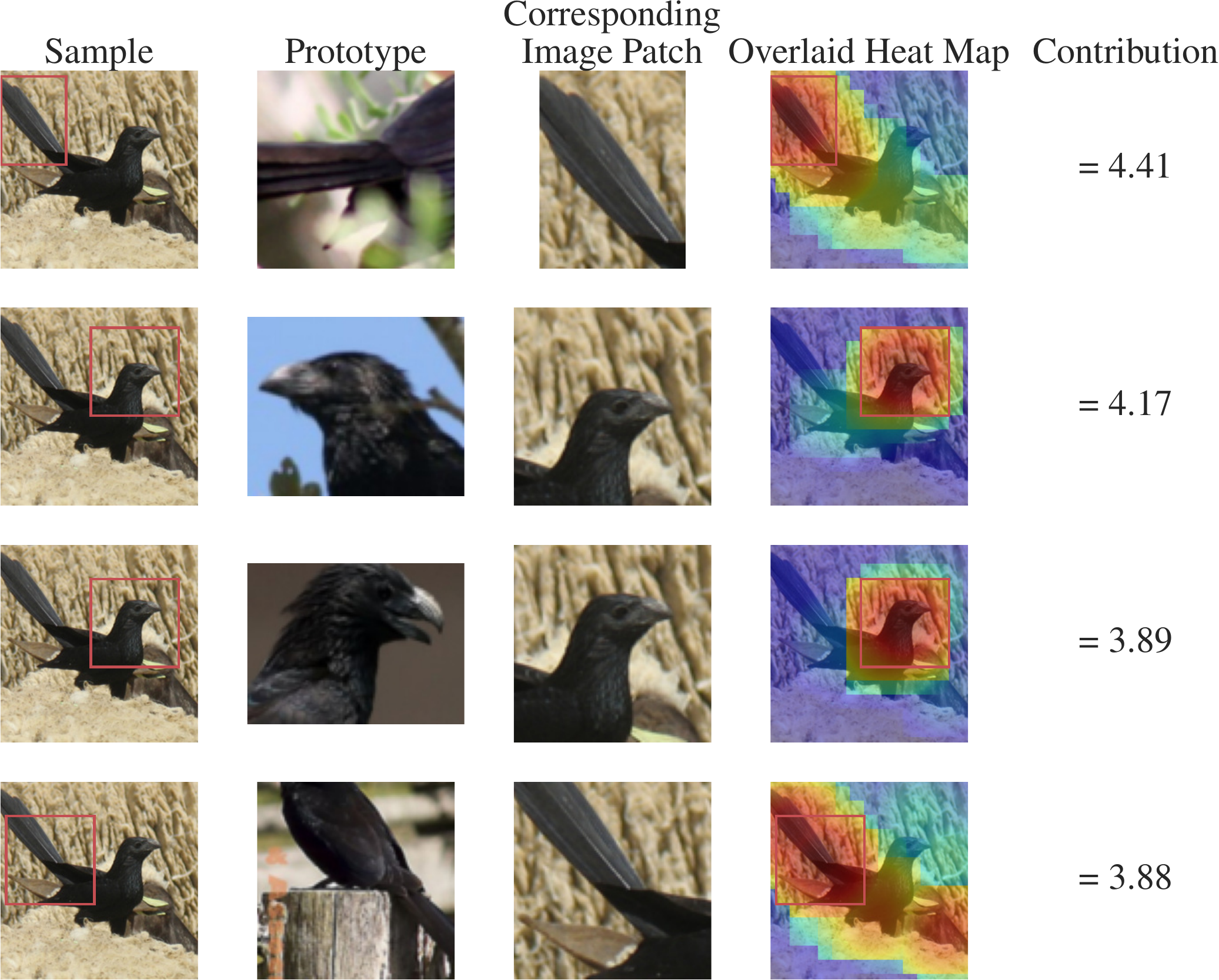}
        \caption{An example of an explanation.}\label{fig:explanation-example}
    \end{subfigure}
    \vspace{-1.5ex}
    \caption{(a) Our proposed architecture, \Ours{}. (b) An example of an explanation with \Ours{} for a Groove-billed Ani. The following are important deviations from \ProtoPNet{}: the backbone $\featuref$ receptive field is constrained, the readout layer $\readoutf$ is simplified, both prototypes and embedded patches truly localize to object parts, and the pixel space mapping is corrected (see Figure~\ref{fig:pixSpaceMap}).}
    \label{fig:ppnet}
    \vspace{-2.5ex}
\end{figure*}

\paragraph{\ProtoPNet{} Architecture}

Here, we go over the \ProtoPNet{} architecture~\cite{protopnet}, a type of \Protonet{}. As much of the formalism overlaps with our approach, Figure~\ref{fig:ppnet} can be referred to for visualization of the architecture.
Let $\dataset {=} 
\{(\sample_1, \ylabel_1), (\sample_2, \ylabel_2), \dots, (\sample_N, \ylabel_N)\}$ 
be the data set
where each sample $\sample_i {\in}  \sR^{3 \times H \times W}$ is an
image with a height of $H$ and a width of $W$,
and each label $\ylabel_i {\in} \{1,\dots,C\}$ represents one of $C$ classes.

A \ProtoPNet{} comprises a neural network backbone responsible for embedding an image. The first component of the backbone is the core $\backbone$, which could be a \ResNet{}~\cite{resnet}, \VGG{}~\cite{vgg}, or \DenseNet{}~\cite{densenet} as in~\cite{protopnet}.
Proceeding, there are the add-on layers $\addonf$ that are responsible for changing the number of channels in the output of $\backbone$. In \ProtoPNet{}, $\addonf$ comprises two $1\times 1$ convolutional layers with ReLU and sigmoid activation functions for the first and second layers, respectively.
The full feature embedding function is denoted by $\featuref = \addonf \circ \backbone$. This function gives us our \textit{embedded patches} $\featuref(\sample_i) = \latentZ_i \in \sR^{D \times H_z \times W_z}$ which have $D$ channels, a height of $H_z$, and a width of $W_z$.

In \ProtoPNet{}, we are interested in finding the most similar embedded patch $\latentz$ for each prototype.
Each prototype can be understood as the embedding of some prototypical part of an object, such as the head of a blue jay as in Figure~\ref{fig:ppnet}.
Each embedded patch can be thought of in the same way -- ultimately, a well-trained network will find that the most similar embedded patch and prototype will both be, \eg{}, the head of a blue jay (\textit{this} prototype looks like \textit{that} embedded patch).
This is accomplished using the prototype layer, $\protof$.
We use the notation $\protof_{\prototype_j}$ to denote the unit that computes the most similar patch $\latentz {\in} \patches(\latentZ_i)$ to prototype $\prototype_j$.
The function $\patches(\latentZ_i)$ yields a set of $D {\times} H_p {\times} W_p$ embedded patches in a sliding window manner
($H_p {=} W_p {=} 1$ in \ProtoPNet{}).
First, the pairwise distances between 
$\patches(\latentZ_i)$
and prototypes $\prototypes {=} \{\prototype_j\}^P_{j=1}$ are computed
using a distance function $\distancef$
where $\prototype_j {\in} \sR^{D {\times} H_p {\times} W_p}$, $H_p$ is the prototype kernel height, $W_p$ is the prototype kernel width, and $P$ is the total number of prototypes.
Each prototype is class-specific and we denote the set of prototypes belonging to class $\ylabel_i$ as $\prototypes_{\ylabel_i} {\subseteq} \prototypes$.
Subsequently, a min-pooling operation is performed to obtain the closest embedded patch for each prototype -- each prototype (\textit{this}) is ``assigned'' a single embedded patch (\textit{that}).
Finally, the distances are converted into similarity scores using a similarity function $\simf$.
Putting this process altogether for unit $\protof_{\prototype_{j}}$, we have
\begin{align}\label{eq:protolayer}
    \protof_{\prototype_{j}}(\latentZ_i) = \simf\Big(
            \min_{\latentz \in \patches(\latentZ_i)} \distancef(\latentz, \prototype_{j})
    \Big)
    .
\end{align} 
We denote the vector of all similarity scores for a sample as $\similarities_i = \protof(\latentZ_i) \in \sR^{P}$.

The architecture ends with a readout layer $\readoutf$ that produces the logits as $\logits_i = \readoutf(\similarities_i)$. In \ProtoPNet{}, $\readoutf$ is a fully-connected layer with positive weights to same-class prototype units and negative weights to non-class prototype units.
Each logit can be interpreted as the sum of similarity scores weighed by their importance to the class of the logit. Note that this readout layer is not reflected in Figure~\ref{fig:ppnet}.
The full \ProtoPNet{} output for a sample is given by $(\readoutf \circ \protof \circ \featuref)(\sample_i)$.

\paragraph{\Protonet{} Desiderata and \ProtoPNet{} Variants}

Many extensions of \ProtoPNet{} have been proposed, some of which make altercations that fundamentally affect the interpretability of the architecture. To differentiate these extensions, we propose a set of desiderata for \Protonets{}:
\begin{enumerate}[nolistsep]  
    \item \textit{Prototypes must correspond directly to image patches}. This can be accomplished via prototype replacement, which grounds prototypes in human-interpretable pixel space (see Section~\ref{sec:fixing-protonets} for details).
    \item \textit{Prototypes must localize to object parts}.
    \item \textit{Case-based reasoning must be describable by linear or simple tree models}.
\end{enumerate}
Architectures that satisfy all three desiderata
are considered to be \textit{3-way \Protonets{}} -- satisfying fewer diminishes the interpretability of the algorithm.

The idea of sharing prototypes between classes has been explored in
\ProtoPShare{}~\cite{ProtoPShare} (prototype merge-pruning) and
\ProtoPool{}~\cite{ProtoPool} (differential prototype assignment).
In \ProtoTree{}~\cite{ProtoTree}, the classification head is replaced by a differentiable tree, also with shared prototypes.
An alternative embedding space is explored in \TesNet{}~\cite{TesNet} based on Grassmann manifolds.
A \Protonet{}-specific knowledge distillation approach is proposed in \ProtoToProto{}~\cite{Proto2Proto} by enforcing that student prototypes and embeddings should be close to those of the teacher.
\DeformableProtoPNet{}~\cite{DeformableProtoPNet} extends the \ProtoPNet{} architecture with deformable prototypes.
\STProtoPNet{}~\cite{ST-ProtoPNet} learns support prototypes that lie near the classification boundary and trivial prototypes that are far from the classification boundary.

In an attempt to improve \ProtoPNet{} visualizations, an extension of layer-wise relevance propagation~\cite{bach2015pixel}, Prototypical Relevance Propagation (\PRP{}), is proposed to create more model-aware explanations~\cite{PRP}. \PRP{} is quantitatively more effective in debugging erroneous prototypes and assigning pixel relevance than the original approach.

\paragraph{\Protonet{}-Like Methods}
The following papers are inspired by \ProtoPNet{} but cannot be considered to be the same class of model. This is due to not fulfilling the proposed \Protonet{} desiderata \#1 (prototypes must correspond directly to image patches) and/or \#3 (case-based reasoning must be describable by linear or simple tree models).

\ViTNeT{}~\cite{ViT-NeT} combines a vision transformer~(ViT) with a neural tree decoder that learns prototypes.
In another transformer-based approach, \ProtoPFormer{}~\cite{ProtoPFormer} exploits the inherent architectural features (local and global branches) of ViTs.
\SemiProtoPNet{}~\cite{Semi-ProtoPNet} fixes the readout weights as \NPProtoPNet{}~\cite{NP-ProtoPNet} does and is used for power distribution network analysis.
In \SDFASAProtoPNet{}~\cite{SDFA-SA-ProtoPNet}, a shallow-deep feature alignment (SDFA) module aligns the similarity structures between deep and shallow layers.
In addition, a score aggregation (SA) module aggregates similarity scores
to avoid learning inter-class information.
Unfortunately, each of these networks omits prototype replacement with the typical justification being that doing so improves task accuracy. In addition, \ViTNeT{} has additional layers after $\protof$
that break the mapping back to pixel space and complicate its case-based reasoning.

\section{The Problem with Existing \Protonets{}}\label{sec:rf_problem}

\begin{figure*}
    \centering
    \begin{subfigure}{0.49\linewidth}
        \centering
        \includegraphics[width=.78\linewidth]{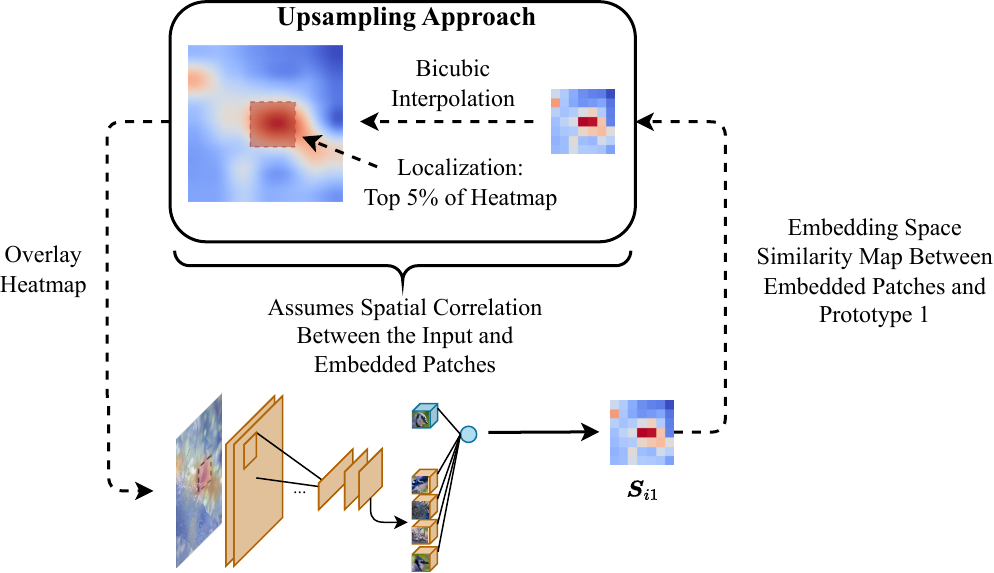}
        \caption{The original upsampling-based pixel space mapping.}\label{fig:viz-mapping-orig}
    \end{subfigure}\hfill%
    \begin{subfigure}{0.49\linewidth}
        \centering
        \includegraphics[width=\linewidth]{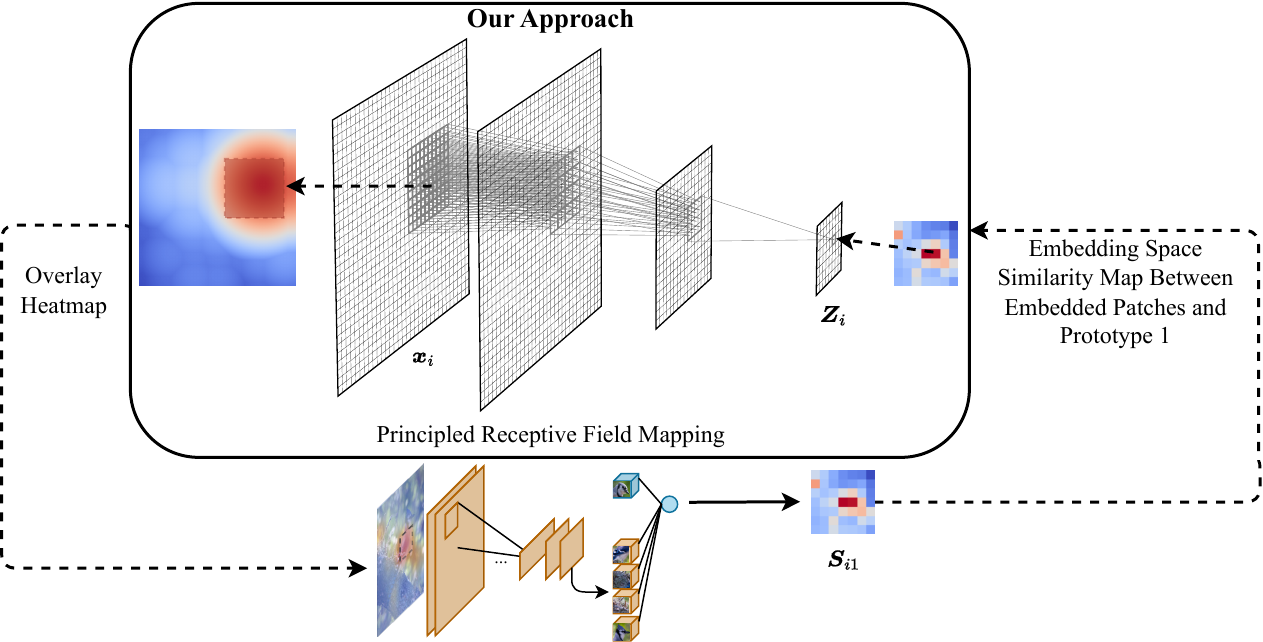}
        \caption{Our proposed receptive field-based pixel space mapping.}\label{fig:viz-mapping-ours}
    \end{subfigure}
    \caption{Visualization of the original and proposed pixel space mapping approaches.}
    \label{fig:pixSpaceMap}
\end{figure*}

Despite the many extensions of \ProtoPNet{}, there are still fundamental issues with object part localization, pixel space grounding, and heat map visualizations, which preclude \textit{any existing \Protonet{} from satisfying all three desiderata} -- all \Protonets{} violate desideratum \#2: prototypes must localize to object parts.
The underlying issues with existing \Protonets{}
arise from 1) their pixel space mapping being reliant on spatial correlation between embedded patches and the input space, which is dubious; 2) their pixel space mapping being receptive field-invariant, arbitrarily localizing to some area in the input.
Rather, intrinsically interpretable models should produce explanations \textit{implicit in the symbolic form of the model itself}~\cite{razIMLForm2022,Schwalbe2023}.

As a refresher, the original visualization process involves three steps. First, a single similarity map $\simmap_{ij} {=} \simmapf_{\prototype_j}(\latentZ_i) \in \sR ^ {H_z / H_p \times W_z / W_p}$ is selected for visualization where $\simmapf_{\prototype_j}$ gives the similarity map for prototype $\prototype_j$. Each element of $\simmap_{ij}$ is given by
%
$
    \simf\left(
        \distancef(\latentz, \prototype_{j})
    \right)
$
%
\noindent
where $\latentz {\in} \patches(\latentZ_i)$.
Subsequently, this map is upsampled from $H_z / H_p {\times} W_z / W_p$ to $H {\times} W$ using bicubic interpolation, producing a heat map $\mM_{ij}{\in} \sR^{H {\times} W}$.
To localize within the image, the smallest bounding box is drawn around the largest 5\% of heat map elements -- this box is of variable size.
While no justification is provided for this approach in the original paper~\cite{protopnet}, 
we believe that
the intuition is that the embedded patches $\latentZ_i$ maintain spatial correlation with the input.
Finally,
$\mM_{ij}$ and the bounding box can be superimposed on the input image for visualization.
From here on out, we will refer to this as the \textit{original pixel space mapping}, which
is visualized in Figure~\ref{fig:viz-mapping-orig}.
It should also be noted that while this pixel space mapping is crucial in establishing interpretability, it is left undiscussed in the vast majority of \ProtoPNet{} extensions.
Immediately, we can see several issues with this approach.

\paragraph{\textit{Here} Does Not Correspond to \textit{There}}
The original pixel space mapping is based on naive upsampling, which is invariant to architectural details.
The approach will always assume that all similarity scores can be mapped to pixel space with a single linear transformation --
an embedded patch at position $\langle t_x, t_y \rangle$
is effectively localized to position $\langle t_x {\cdot} W {\cdot} W_p / W_z, t_y {\cdot} H {\cdot} H_p / H_z \rangle$ in pixel space.
This assumption of spatial correlation from high to low layers is easy to invalidate. For instance, even a simple latent transpose eradicates this correlation.
\textbf{\textit{The similarity scores of embedded patches do not determine where the architecture ``looked'' in the image. Rather, the architecture determines where the similarity scores correspond to in the image.}}
Figure~\ref{fig:thisThatHereThere} demonstrates this discrepancy. Very recently, evidence in \cite{DBLP:conf/iccv/HamdiGG21,sacha2023interpretability} strongly corroborates our arguments about poor localization.
We correct this pixel space mapping according to the receptive fields of the underlying neural architecture.
The original approach also only provides a way to localize a prototype rather than any embedded patch -- our method enables us to do so.
Our approach is described in detail in Section~\ref{sec:fixing-protonets} and we validate its correctness over the original approach in Section~\ref{sec:experiments}.

\paragraph{\textit{This} Does not Correspond to \textit{Just That}}
\ProtoPNet{} and its derivatives all elect to localize to a small region of the input by drawing a bounding box around the largest 5\% of values of heat map $\mM_{ij}$ as shown in Figure~\ref{fig:viz-mapping-orig}.
While this produces alluring visualizations, most of the architectures evaluated in all prior approaches have a mean receptive field of 100\% at the embedding layer\footnote{The lowest mean receptive field of an evaluated architecture is from \textsc{VGG19} ($\sim$70\%)~\cite{protopnet}.}.
\textbf{\textit{A mean receptive field of 100\% means that every element of the embedding layer output is a complex function of every pixel in the input space. Is it fair to say that only $\sim$5\% of the input contributed to some part of a decision?}}
Attribution within the input space spanned by a receptive field is unverifiable from both the feature-selectivity and feature-additivity points of view~\cite{Camburu2019canITrust,falsifiable2020}.
This issue is visualized in Figure~\ref{fig:thisThatHereThere} for an architecture with a mean receptive field under 100\%.
Moreover, while selecting the top 5\% of $\mM_{ij}$ may localize in accordance with its (faulty) intuition, it can actually localize to wildly inaccurate parts of the image (\eg{}, if multiple top values in $\simmap_{ij}$ are all close), breaking the intuition of the (unfaithful) pixel space mapping.
We go on to discuss our solution to this problem in Section~\ref{sec:fixing-protonets}.

\paragraph{The Allure of Visualization}
The original pixel space mapping appears to satisfy human intuitions. However, it is not based on well-justified aspects of explainability.
Beyond the assumption of spatial correlation and naive localization, bicubic interpolation artificially increases the resolution of maps (see Figure~\ref{fig:viz-mapping-orig}), which leads non-experts to believe that per-pixel attributions are estimated.
In our proposed approach, these explanation aspects follow naturally from the
symbolic interpretation of the model itself.

%% file: src/03.Methods.tex
\section{Fixing \Protonets{}}\label{sec:fixing-protonets}

As discussed in Section~\ref{sec:rf_problem}, the underlying issues with \Protonets{} 
arise from 1) the original pixel space mapping being reliant on spatial correlation between embedded patches and the input space, which is dubious; 2) the original pixel space mapping being receptive field-invariant, arbitrarily localizing to some area in the input.
Our proposed architecture, \Ours{} (Pixel-grounded Prototypical part Network), is largely based on \ProtoPNet{} but mitigates these issues through symbolic interpretation of its architecture -- see Figure~\ref{fig:ppnet} for an overview.
In this section, we first describe a new algorithm for the calculation of receptive fields, describe our proposed fixes for prototype visualization and localization, and proceed with additional \Protonet{} corrections and improvements.
With the proposed improvements, \Ours{} is the \textit{only \Protonet{} that truly localizes to object parts}, satisfying all three desiderata.

\paragraph{Receptive Field Calculation Algorithm}

Before delving into our proposed remedies, we describe our approach to computing receptive fields precisely for any architecture.
Our proposed algorithm, \RFAlg{}, takes a neural network as input and outputs the \textit{exact} receptive field of every neuron in the neural network.
Recall that a neuron is a function of a \textit{subset} of pixels defined by its receptive field.
\RFAlg{} represents receptive fields as hypercubes (multidimensional tensor slices).
For instance, the slices for a 2D convolution with a $5{\times} 5$ kernel, stride of 1, and $c_{\text{in}}$ channels at output position $3, 3$ would be $\{\{\llbracket 1,c_{\text{in}} \rrbracket, \llbracket 1,5 \rrbracket,\llbracket 1,5 \rrbracket\}\}$ where $\llbracket a,b \rrbracket$ denotes the slice between $a$ and $b$.
We can compute the \textit{mean receptive field} of a layer as the average number of pixels within the receptive field of each hypercube element of a layer output.
The algorithm does not rely on approximate methods nor architectural alignment assumptions like other approaches~\cite{LuoERF2016,araujo2019computing}.
The full algorithmic details are provided in Appendix C.

\paragraph{Corrected Pixel Space Mapping Algorithm}

\fakeparagraph{From Embedding Space to Pixel Space}
For each prototype $\prototype_j$, we have some $\latentz \in \latentZ_i$ that is most similar. We are interested in knowing where $\latentz$ localizes to in an image $\sample_i$.
With \RFAlg{} applied to the backbone, we have the precise pixel space region that $\latentz$ is a function of -- \textit{this exactly corresponds to that}.
This can also be done for any $\prototype_j$ after prototype replacement.
Additionally, this process can actually be used to visualize any $\latentz \in \latentZ_i$, unlike the procedure specified in the original pixel space mapping~\cite{protopnet}.
See Figure~\ref{fig:viz-mapping-ours} for intuition as to how this process works.

\fakeparagraph{Producing a Pixel Space Heat Map}
In order to compute a pixel space heat map, we propose an algorithm based on \RFAlg{} rather than naively upsampling an embedding space similarity map $\simmap_{ij}$.
Our approach uses the same idea as going from embedding space to pixel space.
Each pixel space heat map $\mM_{ij} \in \sR^{H\times W}$ is initialized to all zeros ($\mathbf{0}^{H \times W}$), and corresponds to a sample $\sample_i$ and a prototype $\prototype_j$.
Let $\mM_{ij}^\emS$ be the region of $\mM_{ij}$ defined by the receptive field of similarity score $\emS \in \simmap_{ij}$.
For each $\emS$, the pixel space heat map is updated as $\mM_{ij}^\emS \gets \max(\mM_{ij}^\emS, \emS)$ where $\max(\cdot)$ is an element-wise maximum that appropriately handles the case of overlapping receptive fields.
We take maxima instead of averaging values due to~\eqref{eq:protolayer}.
Again, see Figure~\ref{fig:viz-mapping-ours} for a visualization of this procedure.
Further algorithmic details are provided in Appendix D.

\paragraph{Improved Localization \& the ``Goldilocks'' Zone}

To reiterate,
the region localized by a \Protonet{} is controlled by the receptive field of the embedding layers of $\featuref$. A fundamental goal of \Protonets{} is to identify and learn prototypical object parts. We propose to achieve this by constraining the receptive field of
$\featuref$
to a range that yields object parts that are both meaningful and interpretable to humans.

\begin{figure}
    \centering
    \includegraphics[width=.75\linewidth]{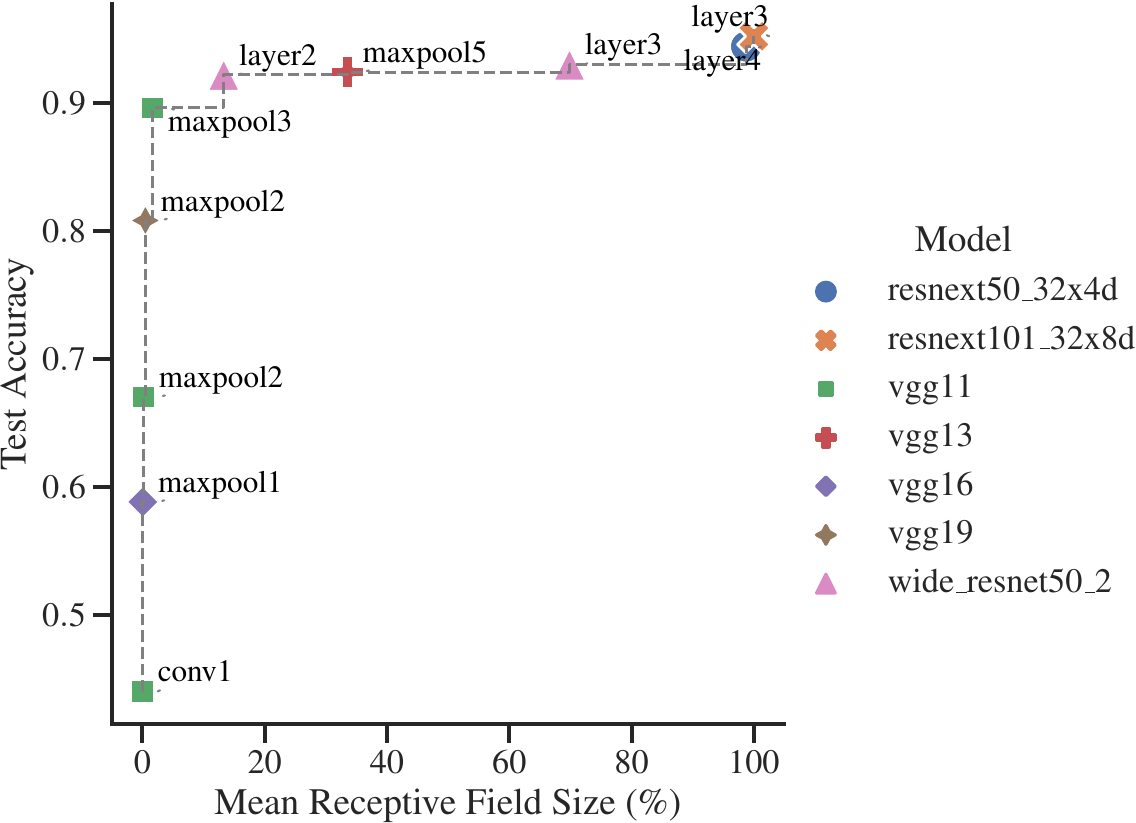}
    \caption{The Pareto front of architectures trained on ImageNette~\cite{imagenette} and evaluated at various intermediate layers. This front details the accuracy-localization size trade-offs and informs backbone selection of \Ours{} as in Section~\ref{sec:experiments}.}
    \label{fig:inet_pareto_front}
\end{figure}

It is well known that the receptive field of a neural network correlates with performance~\cite{LuoERF2016,araujo2019computing} to an extent -- too small or large a receptive field can harm performance due to bias-variance trade-offs~\cite{KoutiniRFReg2019}.
We hypothesize that there is a ``Goldilocks'' zone where the desired receptive field localizes to intelligible object parts without diminishing task performance.
To corroborate this, we evaluate various backbone architectures at intermediate layers on ImageNette~\cite{imagenette}, a subset of ImageNet~\cite{imagenet}.
The evaluation aims to produce architectures suitable for the backbone of \Ours{} according to the criteria outlined prior.
We propose this approach
as performance on subsets of ImageNet has been shown to be reflective of performance on the full dataset~\cite{DBLP:conf/iclr/Dong020}, and ImageNet performance strongly correlates with performance on other vision datasets~\cite{DBLP:conf/cvpr/KornblithSL19}.
We detail the full experiment setup in Appendix F.
The Pareto front of mean receptive field and accuracy for the evaluated architectures is shown in Figure~\ref{fig:inet_pareto_front}.
This front informs our backbone selection as detailed in Section~\ref{sec:experiments}.

\paragraph{Simplified Classification Head}

While the original fully-connected classification head $\readoutf$ is human-interpretable, it has several weaknesses -- its explanation size limits its comprehension~\cite{protoStudy1,protoStudyHIVE} and it requires an additional training stage, adding up to 100 additional epochs in \ProtoPNet{}\footnote{In the original \textsc{ProtoPNet} implementation, as well as subsequent extensions, the last layer is optimized 5 times, each for 20 epochs~\cite{protopnet}.}.
We quantify explanation size in terms of \textit{positive reasoning} and \textit{negative reasoning} about the prediction of a class. For positive reasoning, the number of elements in an explanation with the original fully-connected layer is $2 P / C$: one similarity score per class-specific prototype and a positive weight coefficient. However, considering both positive and negative reasoning
involves $2 P$ total explanation elements.

To address these limitations, we propose to replace the linear layer with a class-wise summation. This operation simply produces the logit of each class as the sum of class-specific similarity scores as
$\hat{\ylabel}_{ic} = \sum_{j : \prototype_j \in \prototypes_{c}} \similarity_{ij}$
where $\hat{\ylabel}_{ic}$ is the logit for class $c$ and $\similarity_{ij}$ is the similarity score for prototype $\prototype_j$.
The layer is visualized in Figure~\ref{fig:ppnet}.
Our new parameter-free readout layer removes the additional training stage and comprises only $P / C$ explanation elements for \textit{both} positive and negative reasoning.
Substituting our layer in the original \ProtoPNet{} configuration for the CUB-200-2011 dataset~\cite{protopnet,cub200} reduces the number of explanation elements for a class prediction from 4,000 down to \textit{just 10}.

\paragraph{Other Improvements}

We also make a few smaller contributions. In prototype replacement, we remove duplicate prototypes (by image or sample) to encourage diversity. If duplicates are found, the next most-similar embedded patch is used in replacement instead.
We also reformulate the similarity function $\simf$ to have lower numerical error (see Appendix G for details) as
$\simf(d) {=} \log(
        \frac
        {1}
        {d {+} \varepsilon} {+} 1
    )$
where $\varepsilon$
mitigates division by zero and the distance $d {=} \distancef(\latentz, \prototype_j)$.
While \ProtoPNet{} uses
$\distancef(\latentz, \prototype_j) {=} \pnorm{2}{\latentz - \prototype_j}^2$,
we elect to use $\distancef(\latentz, \prototype_j) {=} 1 - \frac{\latentz \cdot \prototype}{\pnorm{2}{\latentz} \pnorm{2}{\prototype}}$ (cosine distance), which has a desirable normalizing factor. This distance is also used in~\cite{TesNet,ProtoPMed-EEG,DeformableProtoPNet,XProtoNet}.
In implementation, the distances are computed using generalized convolution~\cite{generalizedConvDist2017,generalizedConv2019,protopnet}.

\input{src/t1.Results.tex}
\vspace{-0.0225in}
\paragraph{Training}
Our multi-stage training procedure is similar to that of \ProtoPNet{}.
The first stage optimizes the full network, except for the readout layer, by 
minimizing \eqref{eq:lossTotal} via stochastic gradient descent
\begin{align}
    \begin{split}\label{eq:lossTotal}
        \frac{1}{N} \hspace{-.5ex} \sum_{i=1}^{N}\hspace{-.2ex}
        \lossXent(\logits_i, \ylabel_i) {+} \lossClsCoef \lossCls(\prototypes\hspace{-.15ex}, \latentZ_i) {+} \lossSepCoef \lossSep(\prototypes\hspace{-.15ex}, \latentZ_i)
    \end{split}
\end{align}
\noindent
where $\lossXent$ is the categorical cross-entropy loss function, $\lossClsCoef$ and $\lossSepCoef$ are auxiliary loss weights, and the auxiliary loss functions, $\lossCls$ and $\lossSep$, are defined as
\begin{align}
    \label{eq:lossCluster}
    \lossCls(\prototypes, \latentZ_i)    &= \frac{1}{N}\sum_{i=1}^N
    \min_{\substack{\prototype_j \in \prototypes_{\ylabel_i} \\ \latentz \in \patches(\latentZ_i)}}
    \distancef(\latentz, \prototype_j) \\
    \label{eq:lossSep}
    \lossSep(\prototypes, \latentZ_i) &= -\frac{1}{N}\sum_{i=1}^N
    \min_{\substack{\prototype_j \notin \prototypes_{\ylabel_i} \\ \latentz \in \patches(\latentZ_i)}}
    \distancef(\latentz, \prototype_j)
    .  
\end{align}
The goal of $\lossCls$ is to ensure that at least one embedded patch of every training image is similar to at least one prototype belonging to the class of the image. In contrast, the goal of $\lossSep$ is to ensure that the embedded patches of every training image are dissimilar from prototypes not belonging to the class of the image.

Subsequently, the prototypes are replaced, which is arguably the most important stage of training
as it grounds prototypes in human-comprehensible pixel space.
The process involves replacing each prototype $\prototype_j$ with an embedded patch $\latentz$ of a training sample of the same class -- the most similar embedded patch replaces the prototype.
In the literature, \textit{prototype replacement} is also referred to as prototype ``pushing'' or ``projection.'' We stick with ``replacement'' for the sake of clarity.
Formally, this update can be written as
    $
    \prototype_j \,{\leftarrow}\, \argmin_{\latentz \in \patches(\latentZ_i)} \distancef(\latentz, \prototype_j), \text{ s.t. } \prototype_j \,{\in}\, \prototypes_{\ylabel_i}
    $%
    . 
\noindent
Without this update, the human interpretation of prototypes is unclear as prototypes are not grounded in pixel space.

In \ProtoPNet{} and its variants, a third stage optimizes the linear readout layer. However, we do not employ this stage as our readout layer is parameter-free.
The multi-stage optimization process can be repeated until convergence.

%% file: src/t1.Results.tex
\begin{table*}
    \footnotesize
    \centering
    \renewcommand{\arraystretch}{1.2}
\addtolength{\tabcolsep}{-1pt}
\begin{tabular}{@{}ccccllrrrrrrrrcc@{}}
\toprule
                                           BBox & D1 & D2 & D3 & Model  &          $f$ & \makecell[r]{Expl.\\Size $+$} & \makecell[r]{Expl.\\Size $\pm$} &   $P$ & MRF &  Acc. &  $\pm$ &  $S_{\text{con}}$ &  $S_{\text{sta}}$ & \makecell{Code\\Avail.} & \makecell{Val.\\Set} \\
\midrule
\multirow{5}{*}{\xmarkInv} &
\multirow{4}{*}{\cmark} & \multirow{4}{*}{\cmark} &\multirow{4}{*}{\cmark}
& \TabOurs{} (Ours) &  ResNeXt{\scriptsize @layer3} &                            \textbf{10} &                            \textbf{10} &  2000 &  100 &   \textbf{81.76} &     0.2 &                56.4 &                \textbf{64.7} &                  \cmark & \cmark           \\
&&& & \TabOurs{} (Ours) &  VGG19{\scriptsize @maxpool5} &                            \textbf{10} &                            \textbf{10} &  2000 &  70.4 &   80.10 &     0.1 &                47.6 &                64.2 &                  \cmark & \cmark           \\
&&&&  \TabOurs{} (Ours) & VGG16{\scriptsize @maxpool5} &                            \textbf{10} &                            \textbf{10} &  2000 &  52.5 &   79.75 &     0.2 &                \textbf{69.5} &                51.6 &                  \cmark & \cmark           \\
&&&&  \TabOurs{} (Ours) & VGG13{\scriptsize @maxpool4} &                            \textbf{10} &                            \textbf{10} &  2000 &  9.69 &   75.32 &     0.2 &                66.9 &                45.0 &                  \cmark & \cmark           \\\cline{2-16}
&\multirow{1}{*}{\cmark} & \multirow{1}{*}{\xmark} &\multirow{1}{*}{\cmark}
& \TabSTProtoPNet{} \cite{ST-ProtoPNet} &  DenseNet161 &                            20 &                            4000 &  2000 &  100 &   80.60 &     -- &                -- &                -- &                  \xmark &                    \bad{?} \\\cline{1-16}
\multirow{7}{*}{\cmarkInv} &
\multirow{7}{*}{\cmark} & \multirow{7}{*}{\xmark} &\multirow{7}{*}{\cmark}&  \TabSTProtoPNet{} \cite{ST-ProtoPNet} &  DenseNet161 &                            20 &                            4000 &  2000 &    100 & \underline{86.10} &    0.2 &                -- &                -- &                  \xmark &                    \bad{?} \\
  &&                                         &        & \TabTesNet{} \cite{TesNet} &  DenseNet121 &                            20 &                            4000 &  2000 &  100 &   84.80 &    0.2 &              63.1 &              \underline{66.1} &                  \cmark &               \xmark \\
    &&                                       &        & \TabProtoPool{} \cite{ProtoPool} &    ResNet152 &                            20 &                             404 &   202 &   100 &  81.50 &    0.1 &              35.7 &              58.4 &                  \cmark &               \xmark \\
      &&                                     &        & \TabProtoPNet{} \cite{protopnet} &  DenseNet121 &                            20 &                            4000 &  2000 &   100 &  80.20 &    0.2 &              24.9 &              58.9 &                  \cmark &               \xmark \\
        &&                                   &        & \TabProtoToProto{} \cite{Proto2Proto} &     ResNet34 &                            20 &                            4000 &  2000 &   100 &  79.89 &     -- &                -- &                -- &                  \cmark &               \xmark \\
          &&                                 &        & \TabProtoPShare{} \cite{ProtoPShare} &  DenseNet161 &                          1200 &                            1200 &   600 &  100 &   76.45 &     -- &                -- &                -- &                  \cmark &               \xmark \\
            &&                               &        & \TabProtoTree{} \cite{ProtoTree,SDFA-SA-ProtoPNet} &  DenseNet121 &                            18 &                             404 &   202 &   100 &  73.20 &     -- &              21.5 &              24.4 &                  \cmark &               \xmark \\
\cmidrule{1-16}\morecmidrules\cmidrule{1-16}
\multirow{2}{*}{\xmarkInv} & 
\multirow{1}{*}{\xmark} & \multirow{1}{*}{\xmark} &\multirow{1}{*}{\cmark} 
                                           & \TabProtoPFormer{} \cite{ProtoPFormer} &       DeiT-S &                            40 &                            8000 &  4000 &  100 &   \textbf{84.85} &     -- &                -- &                -- &                  \cmark &               \xmark \\
& \multirow{1}{*}{\xmark} & \multirow{1}{*}{\xmark} &\multirow{1}{*}{\xmark}
                                           & \TabViTNeT{} \cite{ViT-NeT,ProtoPFormer} &  CaiT-XXS-24 &                             \textbf{8} &                              30 &    15 &   100 &  84.51 &     -- &                -- &                -- &                  \cmark &               \xmark \\\cline{1-16}
\cmarkInv & 
\multirow{1}{*}{\xmark} & \multirow{1}{*}{\xmark} &\multirow{1}{*}{\xmark}
& \TabViTNeT{} \cite{ViT-NeT} &      SwinT-B &                            \underline{10} &                              62 &    31 &   100 &  \underline{91.60} &     -- &                -- &                -- &                  \cmark &               \xmark \\
                                           \cline{1-16}
\bad{?} & 
\multirow{1}{*}{\xmark} & \multirow{1}{*}{\xmark} &\multirow{1}{*}{\cmark}
                                           & SDFA-SA \cite{SDFA-SA-ProtoPNet} &  DenseNet161 &                            20 &                              20 &  2000 &  100 &   86.80 &     -- &              \underline{73.2} &              \underline{73.5} &                  \xmark &                    \bad{?} \\
\bottomrule
\end{tabular}
\addtolength{\tabcolsep}{1pt}
    \caption{\Protonet{} results on CUB-200-2011 with ImageNet used for pre-training. Columns D1, D2, and D3 correspond to the three desiderata established in Section~\ref{sec:xai_methods}. Our approach, \Ours{}, is the only method that is a \textit{3-way \Protonet{}}, satisfying all three desiderata.
    ``BBox'' indicates whether a method crops each image using a bounding box annotation{\protect\footnotemark}.
    The best results of \Protonets{} with and without such annotations are \textbf{bold} and \underline{underlined}, respectively. The table is split based on whether the method meets at least two desiderata.
    The $S_{\text{con}}$ and $S_{\text{sta}}$ scores for other methods are taken from~\cite{SDFA-SA-ProtoPNet} and the top reported accuracy score is taken
    for each method.
    }
    \label{tab:comparativeResults}
\end{table*}

%% file: src/04.Experiments.tex
\section{Experiments \& Discussion}\label{sec:experiments}
To validate our proposed approach, \Ours{}, we
evaluate both its accuracy and interpretability on CUB-200-2011~\cite{cub200}. We also show evaluation results on Stanford Cars~\cite{stanfordcars} in Appendix B.
We draw comparisons against other \Protonets{} with a variety of measures.
We elect to not crop images in CUB-200-2011 by their bounding box annotations to demonstrate the localization capability of \Ours{}.
Hyperparameters, software, hardware, and other reproducibility details are specified in Appendix E.

\footnotetext{A ``BBox'' value of `\bad{?}' means that preprocessing details and code are unavailable.
}

Lastly, upon inspection of the original code base\footnote{\url{https://github.com/cfchen-duke/ProtoPNet}}, we discovered that the test set accuracy is used to influence training of \ProtoPNet{}. 
In fact, neither \ProtoPNet{} nor its extensions for image classification that are mentioned in Section~\ref{sec:xai_methods} employ a validation set in provided implementations.
See Appendix H for further details.

In our implementation, we employ a proper validation set and tune hyperparameters only according to accuracy on this split.

\paragraph{Accuracy}

The experimental results in Table~\ref{tab:comparativeResults}
show that
\Ours{} obtains competitive accuracy with other approaches regardless of whether images are cropped by bird bounding box annotations -- \textit{while we trade off network depth for interpretability, we outperform \ProtoPNet{} and several of its derivatives}.
This is quite favorable as \Ours{} is the only method that truly localizes to object parts.

\paragraph{Interpretability}

We evaluate the interpretability of our approach with several functionally grounded metrics~\cite{doshi2017towards}.
See Figure~\ref{fig:explanation-example} for an example of a \Ours{} explanation.

\fakeparagraph{Relevance Ordering Test~(ROT)}
The ROT is a quantitative measure of how well a pixel space mapping attributes individual pixels according to prototype similarity scores~\cite{PRP}.
First, a pixel space heat map $\mM_{ij}$ is produced for a single sample $\sample_i$ and prototype $\prototype_j$.
Starting from a completely random image, pixels are added back to the random image one at a time in descending order according to $\mM_{ij}$. As each pixel is added back, the similarity score for $\prototype_j$ is evaluated.
This procedure is averaged over each class-specific prototype over 50 random samples.
The faster that the original similarity score is recovered, the better the pixel space mapping is.
Assuming a faithful pixel space mapping, a network with a mean receptive field of, \eg{}, 25\%, will recover the original similarity score after 25\% of the pixels are added back in the worst-case scenario.

We also introduce two aggregate measures of the ROT.
First is the area under the similarity curve (\textbf{AUSC}) which is normalized by the difference between the original similarity score and the baseline value (similarity score for a completely random image)\footnote{AUSC${>}1$ is possible as the maximum possible similarity is unknown.}.
Second is the percentage of pixels added back to recover the original similarity score: pixel percentage to recovery (\textbf{\%2R}).

\input{src/t2.Results-Interp}

We compare our pixel space mapping to the original upsampling approach and \PRP{}~\cite{PRP}. However, the \PRP{} implementation only supports \ResNet{} architectures\footnote{The hard-coded and complex nature of the \textsc{PRP} code base precludes simple extension to other architectures.}, so it is not included in all experiments.
The results in Table~\ref{tab:results-interp} demonstrate that our pixel space mapping best identifies the most important pixels in an image. Naturally, the mean receptive field correlates with both ROT scores.

\fakeparagraph{Explanation Size}
Recall from Section~\ref{sec:fixing-protonets} that the explanation size is the number of elements in an explanation, \ie{}, similarity scores and weight coefficients. This number differs when considering positive or negative reasoning. Due to the original classification head being fully-connected, most \Protonets{} have large explanation sizes when considering both positive and negative reasoning, as shown in Table~\ref{tab:comparativeResults}. In contrast, our explanation size comprises just 10 elements when reasoning about a decision. Our proposed classification head helps to prevent overwhelming users with information, which has been shown to be the case with other \Protonets{}~\cite{protoStudyHIVE}.

\fakeparagraph{Consistency}
The consistency metric~\cite{SDFA-SA-ProtoPNet} quantifies how consistently each prototype localizes to the same human-annotated ground truth part.
It evaluates both semantic similarity quality and the pixel space mapping to a degree.
For a sample $\sample_i$ with label $\ylabel_i$, the pixel space mapping is computed for each prototype $\prototype_j \in \prototypes_{\ylabel_i}$. Let $o_{\prototype_j}(\sample_i) \in \sR^{K}$ be a binary vector indicating which of $K$ object parts are contained within the region localized by the pixel space mapping. Let $u(\sample_i) \in \sR^{K}$ be a binary vector indicating which of the $K$ object parts are actually visible in $\sample_i$.
A single object part is associated with $\prototype_j$ by taking the maximum frequency of an object part present in the pixel space mapping region across all applicable images.
A prototype is said to be consistent if this frequency is at least $\mu$, \ie{},
\begin{align*}
    S_{\text{con}} = \frac{1}{P} \sum_{j=1}^P \mathbbm{1}\Bigg[ \max\Bigg(\sum_{\sample_i \in \samples_{c_j}} \left( o_{\prototype_j}(\sample_i) \oslash u(\sample_i) \right)\Bigg) \ge \mu \Bigg]
\end{align*}
where $\samples_{c_j}$ are samples of the same class allocated to $\prototype_j$, $\oslash$ denotes element-wise division, and $\mathbbm{1}$ is the indicator function.
To compare with results reported in~\cite{SDFA-SA-ProtoPNet}, we change the receptive field size in our pixel space mapping to equal this, as well as set $\mu = 0.8$.
A notable weakness of the evaluation approach is that it uses a fixed $72 \times 72$ pixel region independent of the architecture.
While the approach is not perfect, it allows for reproducible and comparative interpretability evaluation between \Protonet{} variants.

Results are shown in Tables~\ref{tab:comparativeResults} and \ref{tab:results-interp} for CUB-200-2011, which provides human-annotated object part annotations. We outperform \ProtoPNet{} and many of its variants, as well as the original pixel space mapping (Table~\ref{tab:results-interp}).

\fakeparagraph{Stability}
The stability metric~\cite{SDFA-SA-ProtoPNet} measures how robust object part association is when noise is added to an image.
Simply, some noise $\epsilon \sim \mathcal{N}(0, \sigma ^2)$ is added to each sample $\sample_i$ and the object part associations are compared as
\begin{align*}
    S_{\text{sta}} = \frac{1}{P} \sum_{j=1}^P \frac{\sum_{\sample_i \in \samples_{c_j}}\mathbbm{1}\left[ o_{\prototype_j}(\sample_i) = o_{\prototype_j}(\sample_i + \epsilon) \right]}
    {|\samples_{c_j}|}
    .
\end{align*}

Following~\cite{SDFA-SA-ProtoPNet}, we set $\sigma {=} 0.2$.
Results in Tables~\ref{tab:comparativeResults} and \ref{tab:results-interp} support the robustness of
\Ours{}
compared to other \Protonets{} and the original pixel space mapping.
There is a marginal decrease in stability as the receptive field lessens.

%% file: src/t2.Results-Interp.tex
\begin{table}[t]
    \footnotesize
    \centering
    \renewcommand{\arraystretch}{1.2}
\begin{tabular}{@{}lrrlrrrr@{}}
\toprule
Backbone & MRF & \makecell[r]{Acc.\\↑} & PSM        &  \makecell[r]{$S_{\text{con}}$\\↑} &  \makecell[r]{$S_{\text{sta}}$\\↑} &  \makecell[r]{AUSC\\↑} &  \makecell[r]{\%2R\\↓} \\
\midrule
\multirow{2}{*}{\makecell[l]{VGG11\\{\scriptsize @maxpool4}}} & \multirow{2}{*}{8.31} & \multirow{2}{*}{72.9} & Ours &                               \textbf{65.3} &                               \textbf{48.3} &                   \textbf{0.99} &                   \textbf{11.2} \\
                                              &      &      & Orig. &                               45.8 &                               44.0 &                   0.90 &                   30.5 \\
\cline{1-8}
\cline{2-8}
\cline{3-8}
\multirow{2}{*}{\makecell[l]{VGG13\\{\scriptsize @maxpool4}}} & \multirow{2}{*}{9.69} & \multirow{2}{*}{75.3} & Ours &                               \textbf{66.9} &                               \textbf{45.0} &                  \textbf{0.97} &                   \textbf{13.0} \\
                                              &      &      & Orig. &                               48.1 &                               41.8 &                   0.88 &                   84.1 \\
\cline{1-8}
\cline{2-8}
\cline{3-8}
\multirow{2}{*}{\makecell[l]{VGG16\\{\scriptsize @maxpool4}}} & \multirow{2}{*}{15.7} & \multirow{2}{*}{76.4} & Ours &                               \textbf{62.0} &                               \textbf{46.4} &                   \textbf{1.02} &                   \textbf{6.98} \\
                                              &      &      & Orig. &                               46.8 &                               42.2 &                   0.89 &                   35.5 \\
\cline{1-8}
\cline{2-8}
\cline{3-8}
\multirow{2}{*}{\makecell[l]{VGG19\\{\scriptsize @maxpool4}}} & \multirow{2}{*}{22.8} & \multirow{2}{*}{77.1} & Ours &                              \textbf{ 60.1} &                               \textbf{42.5} &                   \textbf{0.94} &                  \textbf{ 21.4} \\
                                              &      &      & Orig. &                               48.4 &                               41.3 &                   0.80 &                   99.9 \\
\cline{1-8}
\cline{2-8}
\cline{3-8}
\multirow{2}{*}{\makecell[l]{VGG13\\{\scriptsize @maxpool5}}} & \multirow{2}{*}{33.5} & \multirow{2}{*}{78.1} & Ours &                               \textbf{67.0 }&                              \textbf{ 42.5} &                  \textbf{ 0.90 }&                 \textbf{  29.5} \\
                                              &      &      & Orig. &                               43.7 &                               39.9 &                   0.81 &                   99.2 \\
\cline{1-8}
\cline{2-8}
\cline{3-8}
\multirow{2}{*}{\makecell[l]{VGG16\\{\scriptsize @maxpool5}}} & \multirow{2}{*}{52.5} & \multirow{2}{*}{79.8} & Ours &                               \textbf{69.5} &                              \textbf{ 51.6} &                   \textbf{0.90} &                   \textbf{32.0} \\
                                              &      &      & Orig. &                               44.1 &                               42.4 &                   0.82 &                   55.5 \\
\cline{1-8}
\cline{2-8}
\cline{3-8}
\multirow{2}{*}{\makecell[l]{WRN50\\{\scriptsize @layer3}}} & \multirow{2}{*}{69.9} & \multirow{2}{*}{80.1} & Ours &                               \textbf{56.4} &                              \textbf{ 64.7} &                   \textbf{0.93 }&                  \textbf{ 13.0} \\
                                              &      &      & Orig. &                               56.4 &                               47.6 &                   0.85 &                   39.6 \\
\cline{1-8}
\cline{2-8}
\cline{3-8}
\multirow{2}{*}{\makecell[l]{VGG19\\{\scriptsize @maxpool5}}} & \multirow{2}{*}{70.4} & \multirow{2}{*}{80.1} & Ours &                               \textbf{47.6} &                               \textbf{64.2 }&                   \textbf{0.92} &                   \textbf{43.4} \\
                                              &      &      & Orig. &                               45.8 &                               46.0 &                   0.85 &                   92.9 \\
\cline{1-8}
\cline{2-8}
\cline{3-8}
\multirow{3}{*}{\makecell[l]{ResNet18\\{\scriptsize @layer2}}} & \multirow{3}{*}{15.4} & \multirow{3}{*}{57.2} & Ours &                               \textbf{59.2} &                               \textbf{46.6} &                  \textbf{ 0.98} &                   \textbf{4.10} \\
                                              &      &      & Orig. &                               25.2 &                               45.6 &                   0.88 &                   96.8 \\
                                              &      &      & \TabPRP{} &                                 -- &                                 -- &                   0.95 &                   25.4 \\
\cline{1-8}
\cline{2-8}
\cline{3-8}
\multirow{3}{*}{\makecell[l]{ResNet50\\{\scriptsize @layer3}}} & \multirow{3}{*}{69.8} & \multirow{3}{*}{76.6} & Ours &                               47.9 &                               \textbf{62.0} &                  \textbf{ 0.58} &                   \textbf{72.8} \\
                                              &      &      & Orig. &                               \textbf{53.5} &                               42.7 &                   0.42 &                   97.8 \\
                                              &      &      & \TabPRP{} &                                 -- &                                 -- &                   0.34 &                  100.0 \\
\bottomrule
\end{tabular}
\vspace{-1ex}
    \caption{Evaluation of pixel space mapping (PSM) methods with functionally-grounded interpretability metrics. Methods are compared on \Ours{} with ``Goldilocks'' zone and \ResNet{} backbones
    on CUB-200-2011 (no BBox cropping). Our PSM outperforms
    both the original and \PRP{}
    PSMs across \textit{all} backbones.}
    \label{tab:results-interp}
    \vspace{-2ex}
\end{table}

%% file: src/05.Discussion.tex
\section{Limitations and Future Work}
The receptive field constraint is a design choice and is inherently application-specific, subject to data characteristics and interpretability requirements. Future work should investigate multi-scale receptive fields and automated receptive field design techniques.
Nevertheless, we \textit{trade off network depth for significant gains in interpretability with very little penalty in accuracy}.
Prior studies have shown that \Protonets{} have a semantic similarity gap with humans, prototypes can be redundant or indistinct, and limited utility in improving human performance~\cite{DBLP:conf/iccv/HamdiGG21,protoAttack,protoStudyHIVE,protoStudy1}.
Moreover, the consistency and stability evaluation metrics are imperfect.
Although we
improve upon interpretability over other networks, human studies are needed to understand other facets of interpretability, such as trustworthiness, acceptance, and utility~\cite{Schwalbe2023}.
In the future, architectural improvements should be made, \eg{}, the enriched embedding space of \TesNet{}, prototype diversity constraints~\cite{ProtoSeg,DPNet,ST-ProtoPNet}, and human-in-the-loop training~\cite{ProSeNet}.

%% file: src/suppmat.tex
\renewcommand\thefigure{\thesection\arabic{figure}}
\renewcommand\thetable{\thesection\arabic{table}}

\renewcommand{\contentsname}{Supplemental Material Contents}
\tableofcontents

\section{Symbols and Functions}

Tables~\ref{tab:functions} and \ref{tab:symbols} describe the functions and symbols used in this paper, respectively.

\begin{table}
    \footnotesize
    \centering
    \begin{tabular}{lr}
        \toprule
        Function & Description \\
        \midrule
        $\distancef$ & The distance function \\
        $\backbone$ & The core neural network backbone \\
        $\addonf$ & The add-on layers to the backbone \\
        $\featuref$ & The feature encoding function \\
        $\protof$ & The prototype layer \\
        $\readoutf$ & The readout layer \\
        $\simf$ & The similarity function \\
        $\simmapf$ & The similarity map function \\
        $\texttt{patches}$ & Yields patches from an embedded image \\
        $\lossTotal$ & The total loss function \\
        $\lossXent$ & The cross-entropy loss function \\
        $\lossCls$ & The cluster loss function \\
        $\lossSep$ & The separation loss function \\
        $\lossReadout$ & The readout loss function \\
        \bottomrule
    \end{tabular}
    \caption{Functions used in this paper.}
    \label{tab:functions}
\end{table}

\begin{table}
    \footnotesize
    \centering
    \begin{tabular}{lcr}
        \toprule
        Symbol & Shape & Description \\
        \midrule
        $D$ & -- & The prototype dimensionality \\
        $P$ & -- & The number of prototypes \\
        $N$ & -- & The number of samples \\
        $C$ & -- & The number of classes \\
        $H$ & -- & The height of a sample \\
        $W$ & -- & The width of a sample \\
        $H_z$ & -- & The height of $\latentz$ \\
        $W_z$ & -- & The width of $\latentz$ \\
        $H_p$ & -- & The height of a prototype \\
        $W_p$ & -- & The width of a prototype \\
        $\prototype$ & $D \times H_p \times W_p$ & A prototype \\
        $\prototypes$ & $P \times D \times H_p \times W_p$ & The tensor of all prototypes \\
        $\latentz$ & $D \times H_p \times W_p$ & A single patch (embedded vector) \\
        $\latentZ$ & $D \times H_z \times W_z$ & A tensor of embeddings \\
        $\similarities$ & $P$ & A set of similarity scores \\
        $\simmap$ & $H_z / H_p \times W_z / W_p$ & A similarity map for one prototype \\
        $\dataset$ & $N \times 3 \times H \times W$ & A dataset \\
        $\samples$ & $N \times 3 \times H \times W$ & All samples in a dataset \\
        $\sample$ & $3 \times H \times W$ & A sample \\
        $\ylabels$ & -- & All ground truth labels \\
        $\ylabel$ & -- & The ground truth label \\
        $\logits$ & $C$ & The predicted logits \\
        $\prediction$ & -- & The prediction \\
        $\lossClsCoef$ & -- & Loss function coefficient for $\lossCls$ \\
        $\lossSepCoef$ & -- & Loss function coefficient for $\lossSep$ \\
        $\lossReadoutCoef$ & -- & Loss function coefficient for $\lossReadout$ \\
        $\readoutW$ & $P \times C$ & The weight matrix of $\readoutf$ \\
        $\readoutw$ & -- & An element of $\readoutW$ \\
        \bottomrule
    \end{tabular}
    \caption{Symbols used in this paper.}
    \label{tab:symbols}
\end{table}

\section{More Results}

\begin{figure}
    \centering
    \includegraphics[width=\linewidth]{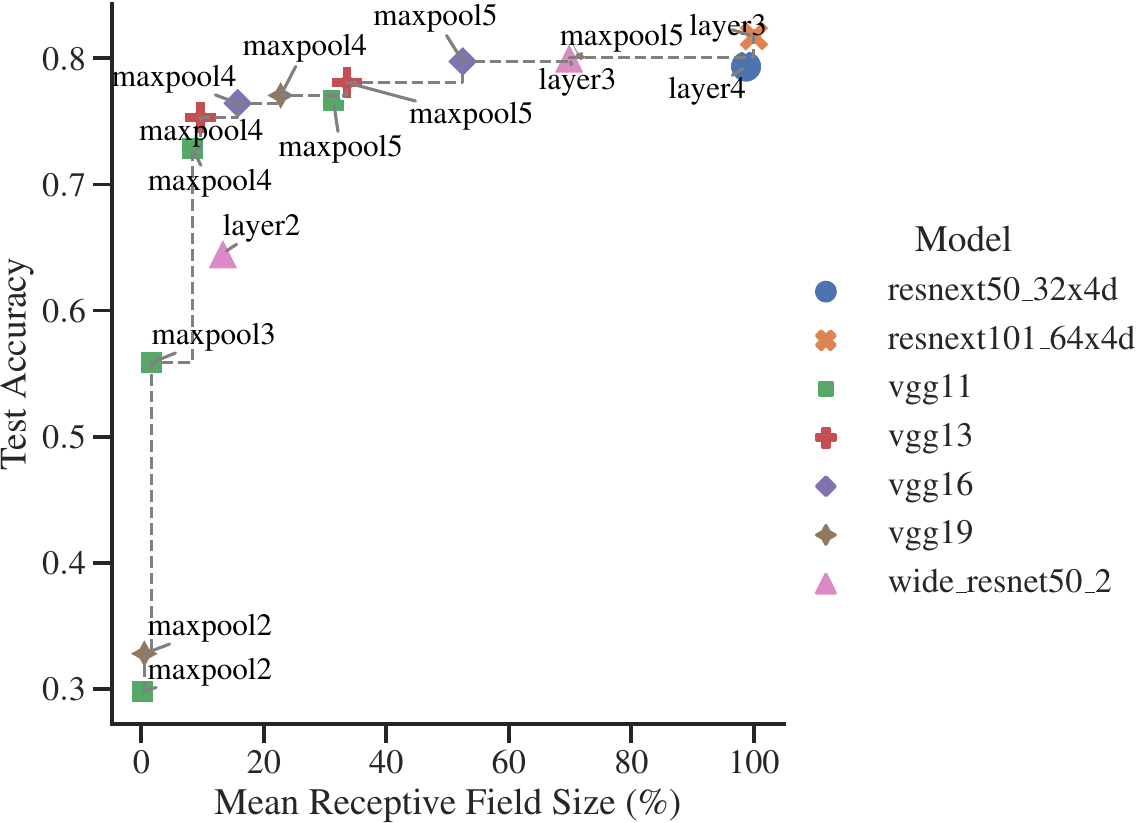}
    \caption{Our approach, \Ours{}, evaluation on CUB-200-2011 with various backbones discovered by the ImageNette proxy approach outlined in
    the main text. The Pareto front is given by the dashed line.}
    \label{fig:pareto-cub}
\end{figure}

\begin{figure}
    \centering
    \includegraphics[width=\linewidth]{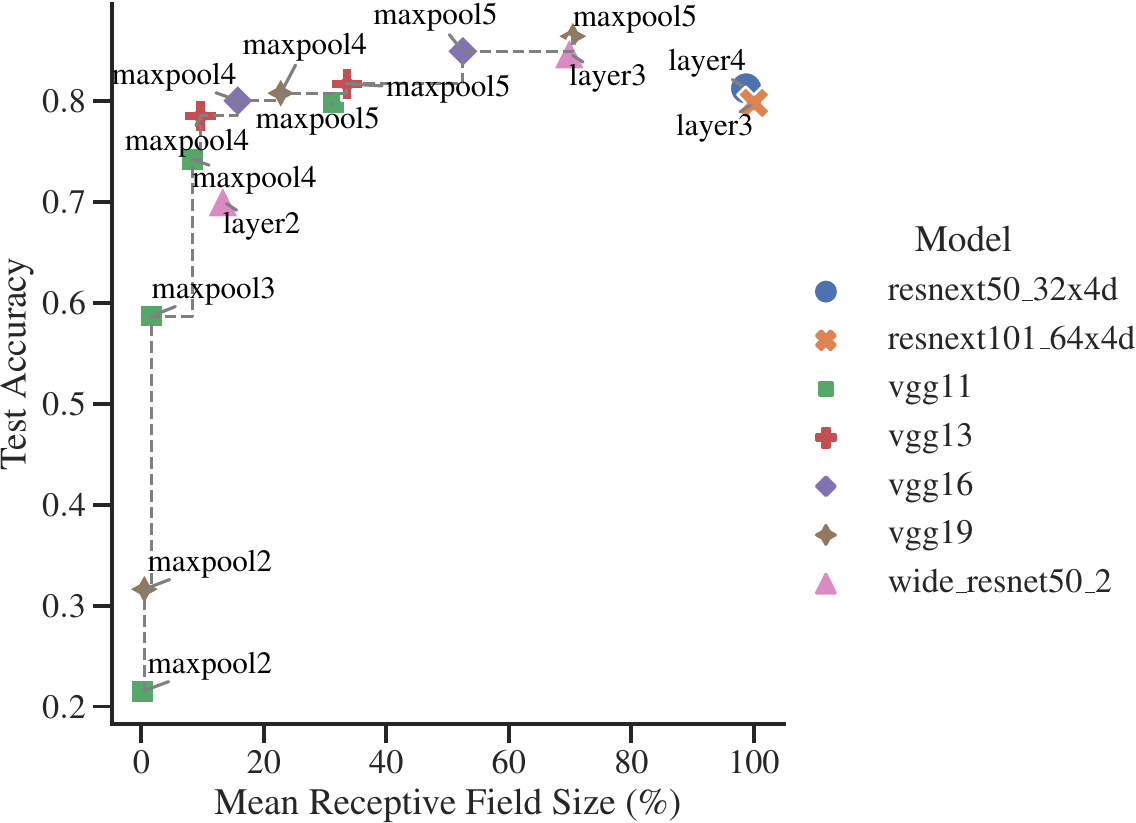}
    \caption{Our approach, \Ours{}, evaluation on Stanford Cars with various backbones discovered by the ImageNette proxy approach outlined in
    the main text. The Pareto front is given by the dashed line.}
    \label{fig:pareto-cars}
\end{figure}

\begin{figure*}
    \centering
    \begin{subfigure}[b]{.4\linewidth}%
        \centering
        \includegraphics[width=\linewidth]{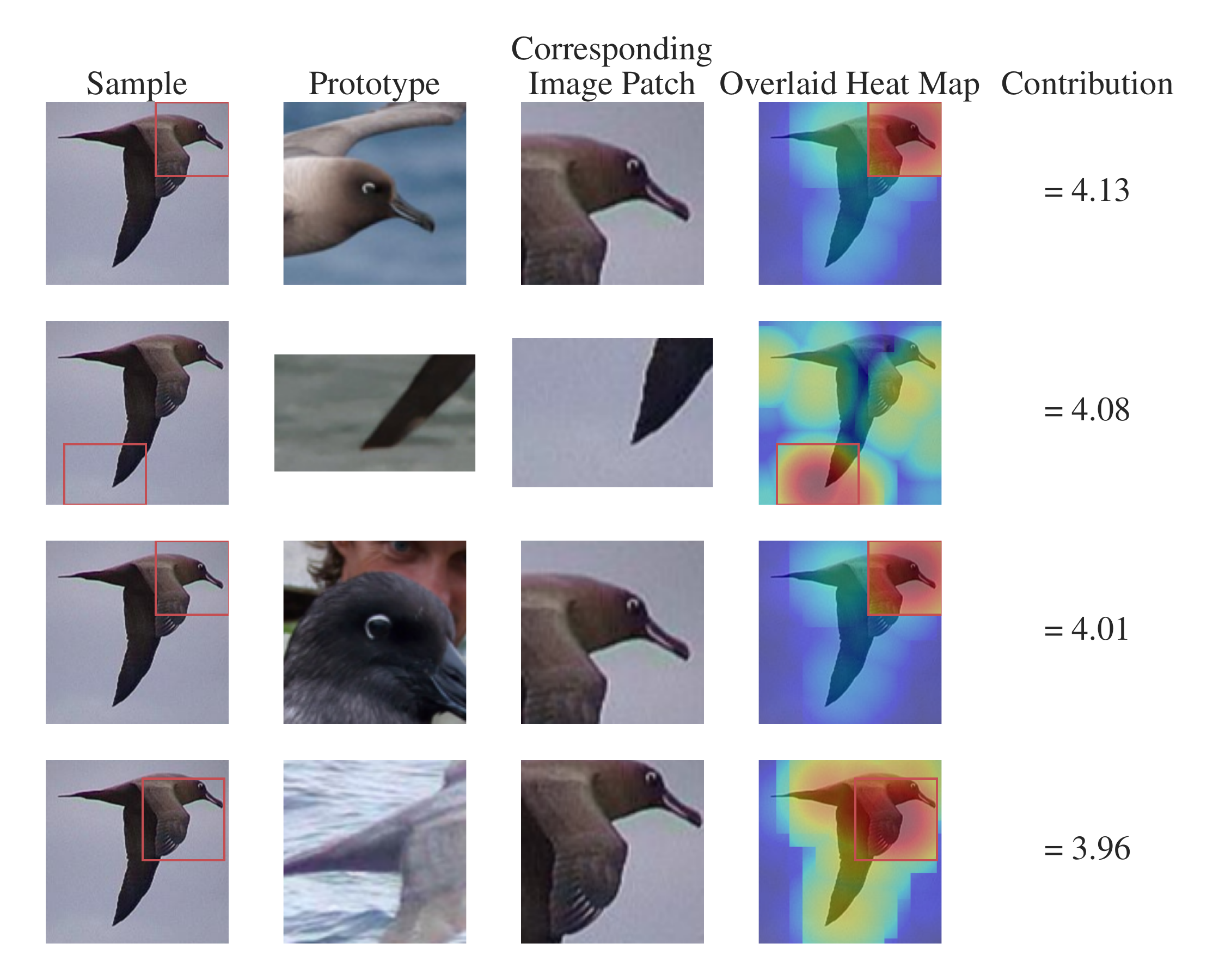}
    \end{subfigure}%
    \begin{subfigure}[b]{.4\linewidth}%
        \centering
        \includegraphics[width=\linewidth]{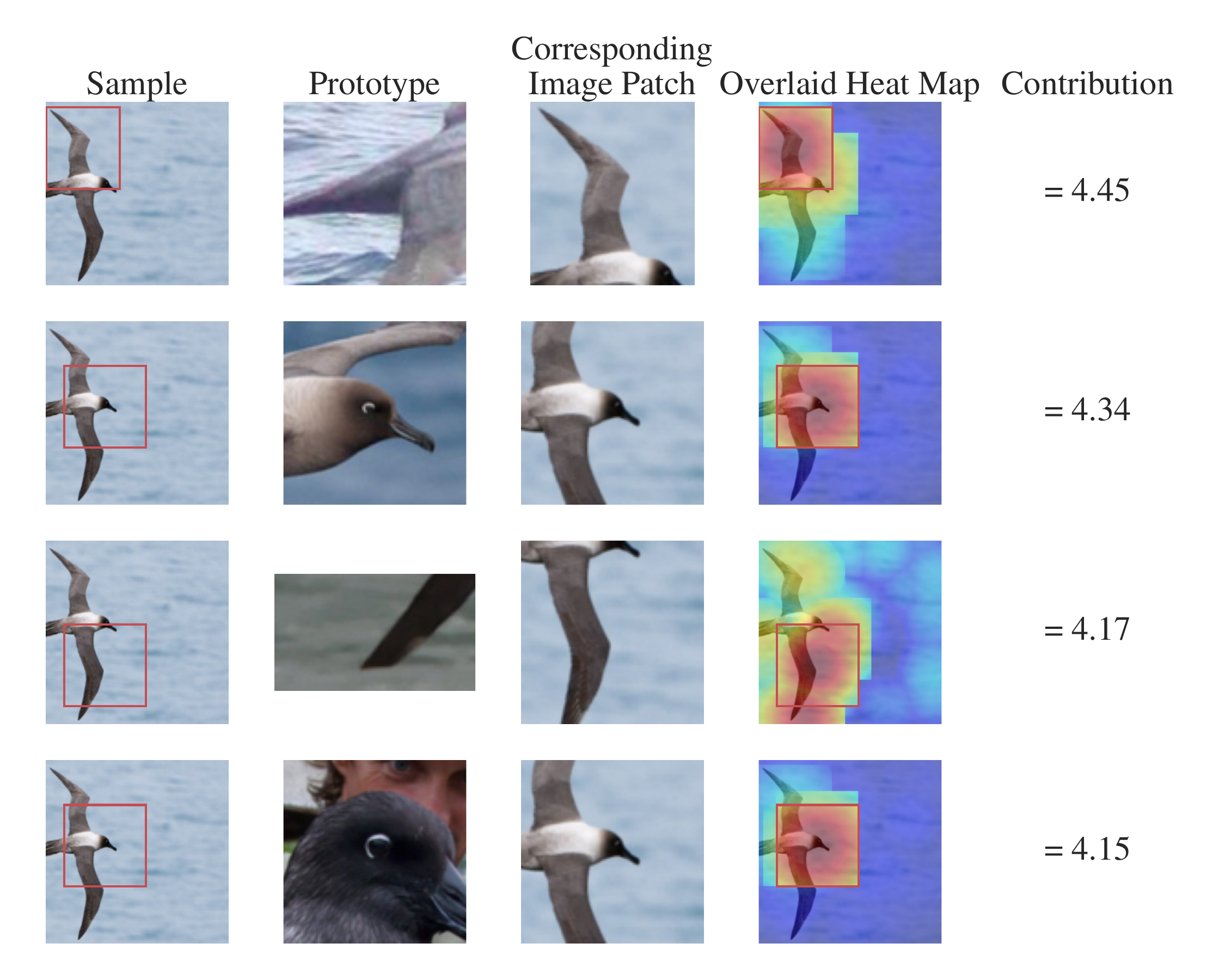}
    \end{subfigure}\\%
    \begin{subfigure}[b]{.4\linewidth}%
        \centering
        \includegraphics[width=\linewidth]{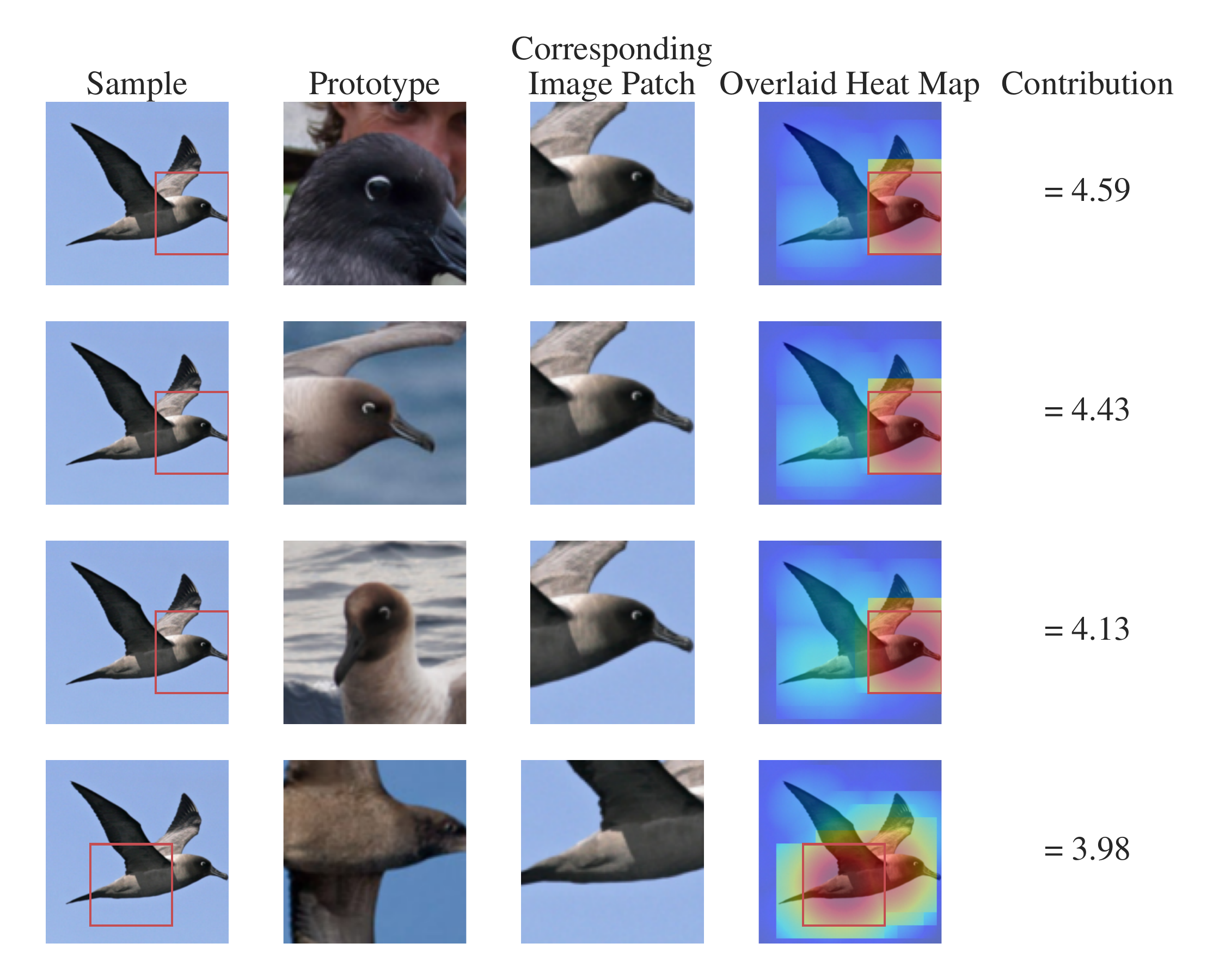}
    \end{subfigure}%
    \begin{subfigure}[b]{.4\linewidth}%
        \centering
        \includegraphics[width=\linewidth]{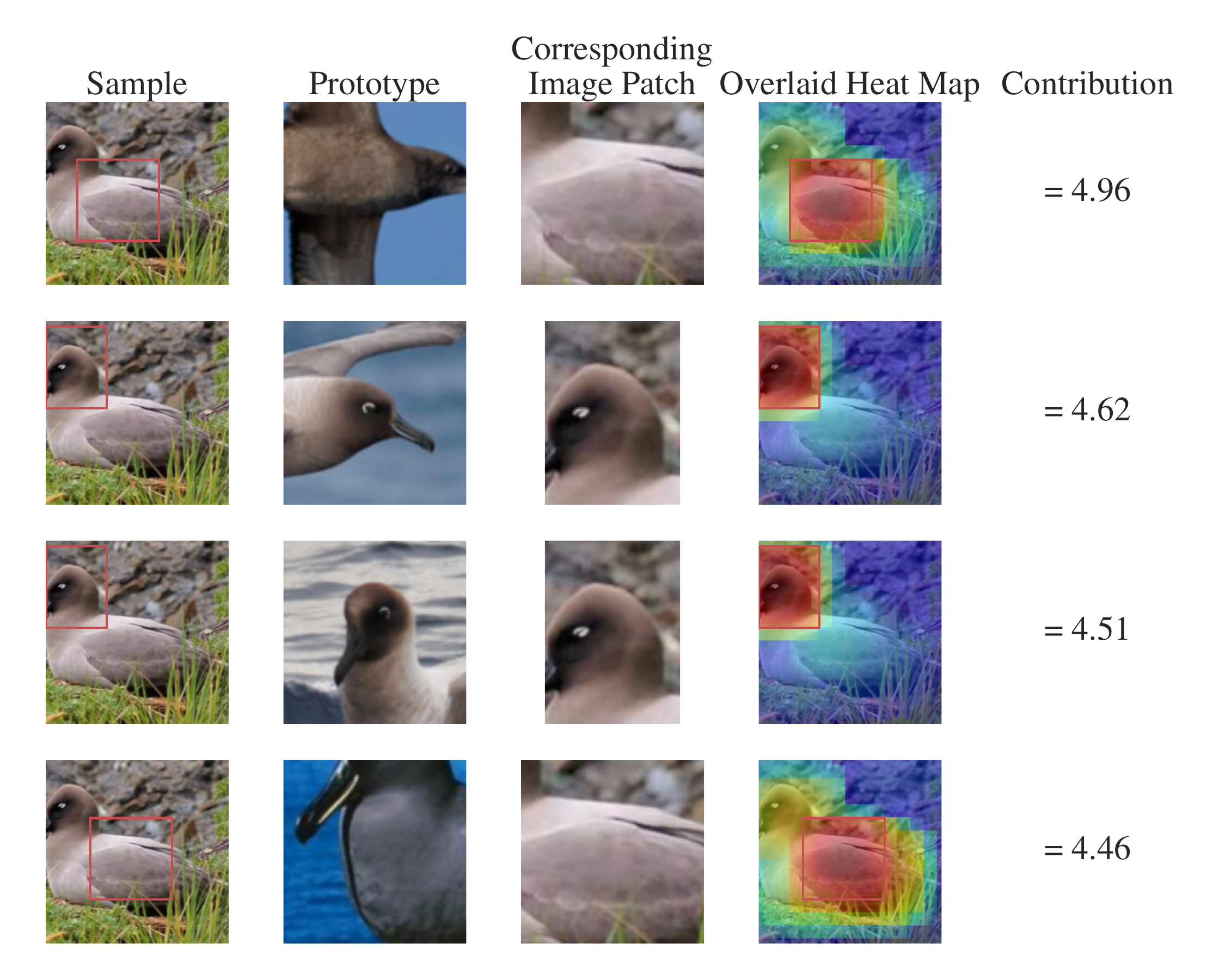}
    \end{subfigure}\\%
    \begin{subfigure}[b]{.4\linewidth}%
        \centering
        \includegraphics[width=\linewidth]{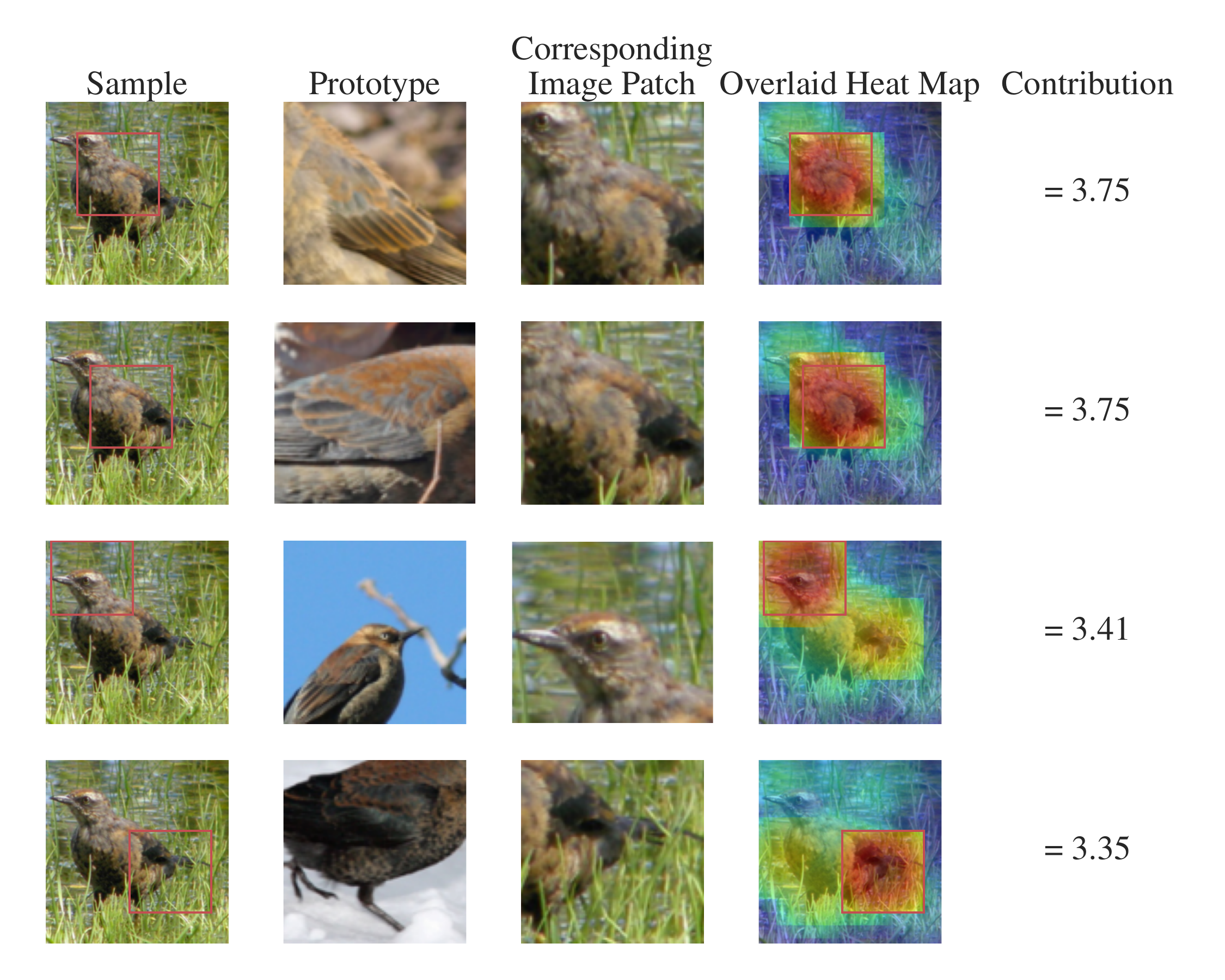}
    \end{subfigure}%
    \begin{subfigure}[b]{.4\linewidth}%
        \centering
        \includegraphics[width=\linewidth]{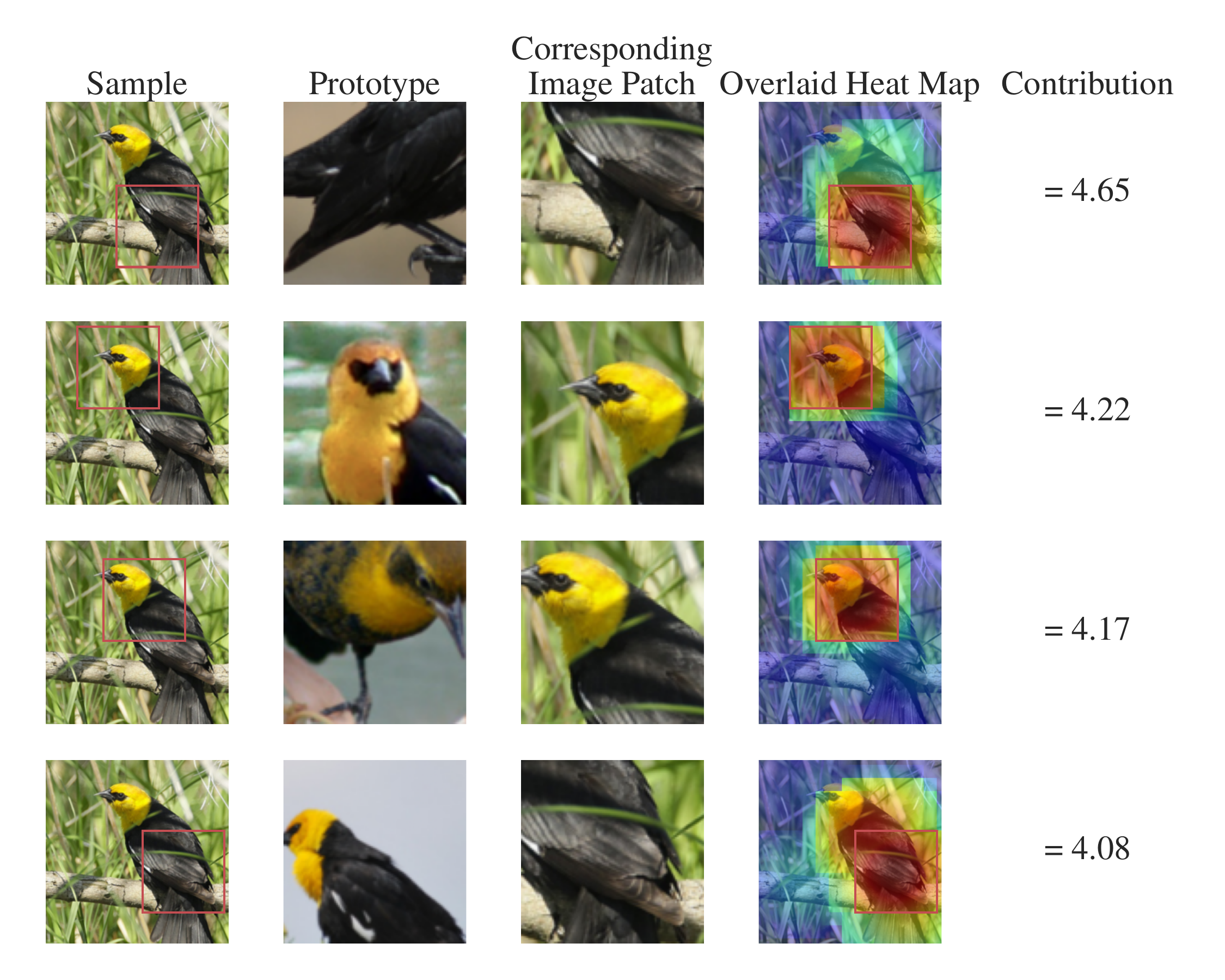}
    \end{subfigure}\\%
    \begin{subfigure}[b]{.4\linewidth}%
        \centering
        \includegraphics[width=\linewidth]{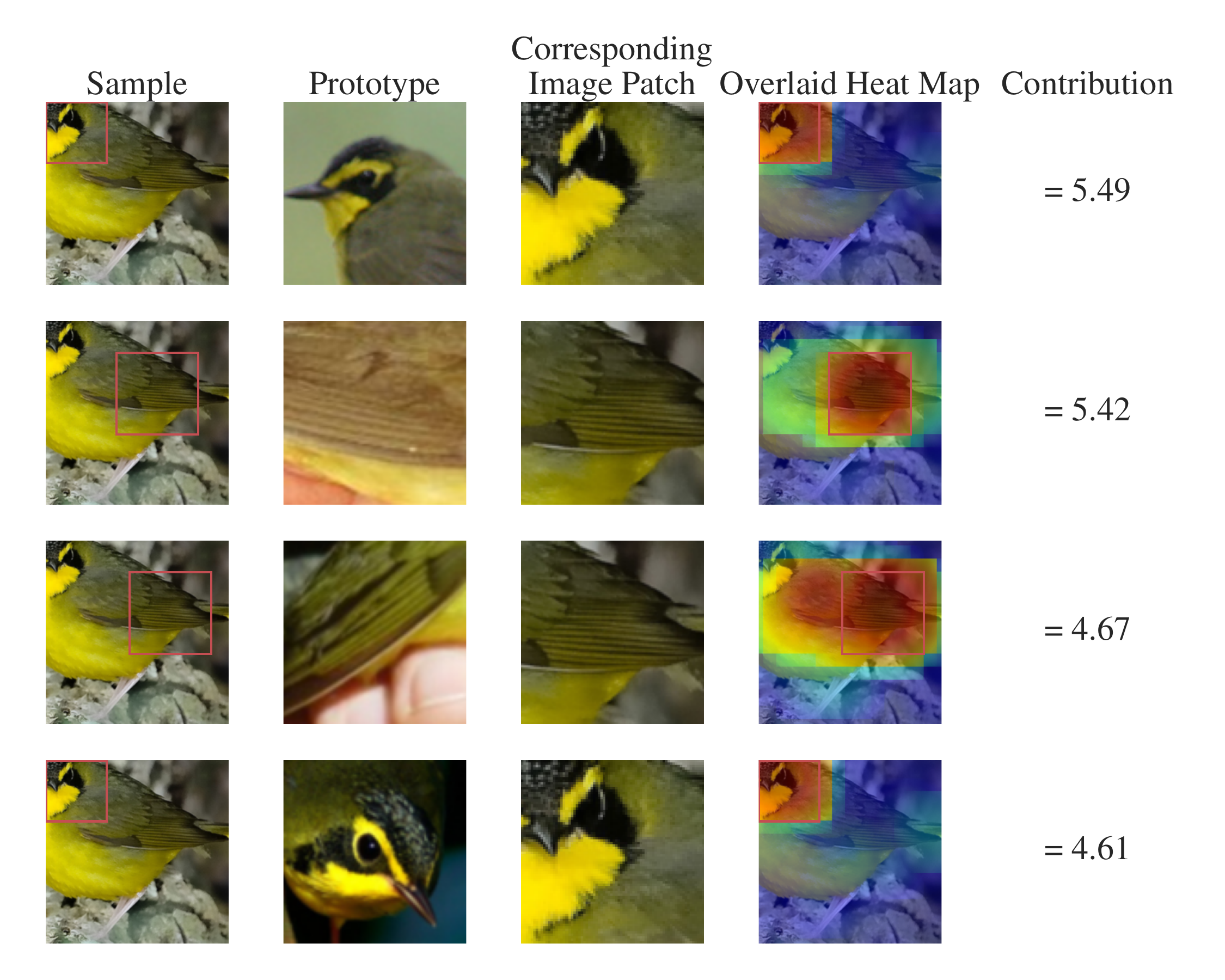}
    \end{subfigure}%
    \begin{subfigure}[b]{.4\linewidth}%
        \centering
        \includegraphics[width=\linewidth]{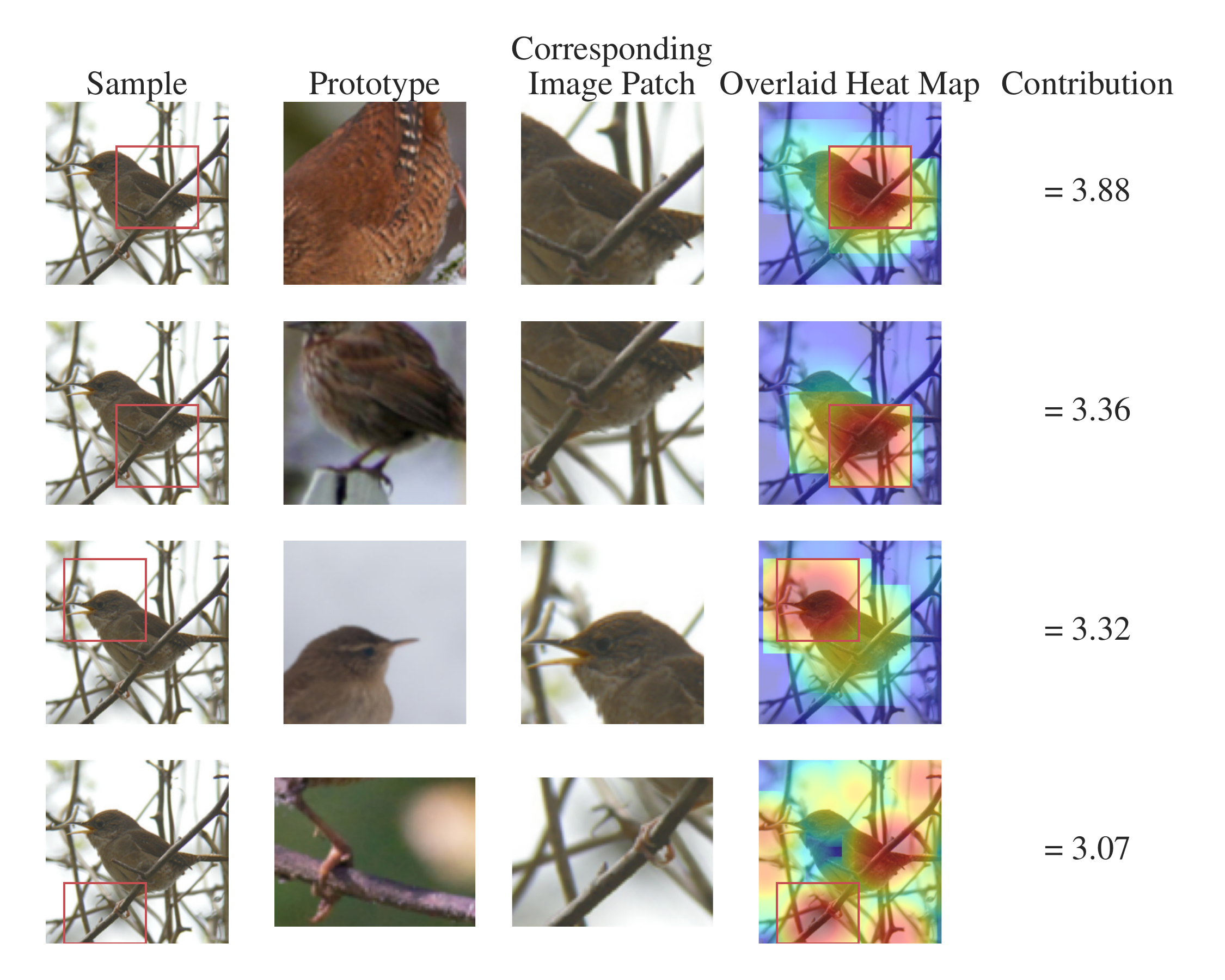}
    \end{subfigure}%
    \vspace{-2ex}
    \caption{Examples of \Ours{} explanations on CUB-200-2011.}
    \label{fig:more-explanations}
\end{figure*}

\begin{figure*}
    \centering
    \begin{subfigure}[b]{.4\linewidth}%
        \centering
        \includegraphics[width=\linewidth]{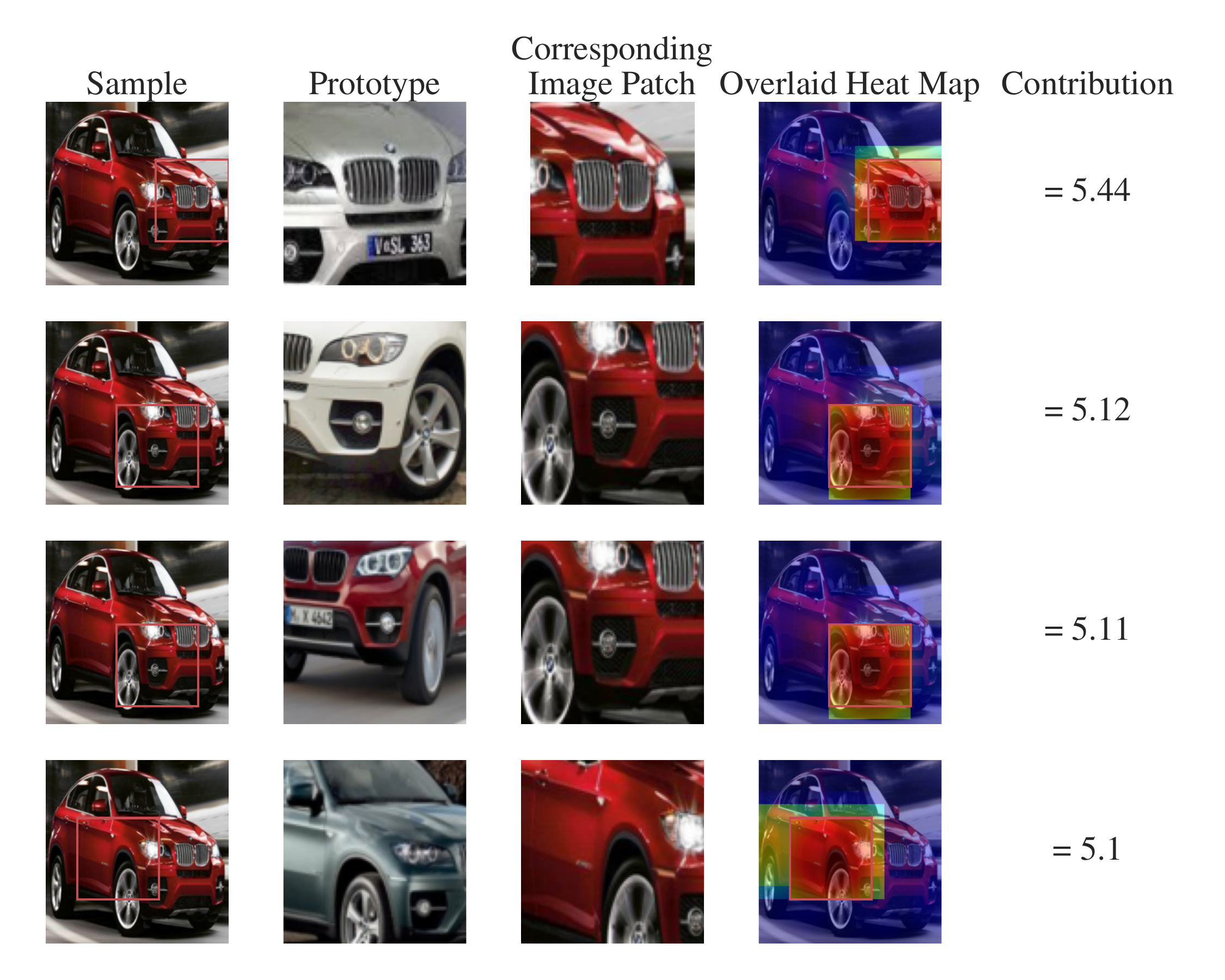}
    \end{subfigure}%
    \begin{subfigure}[b]{.4\linewidth}%
        \centering
        \includegraphics[width=\linewidth]{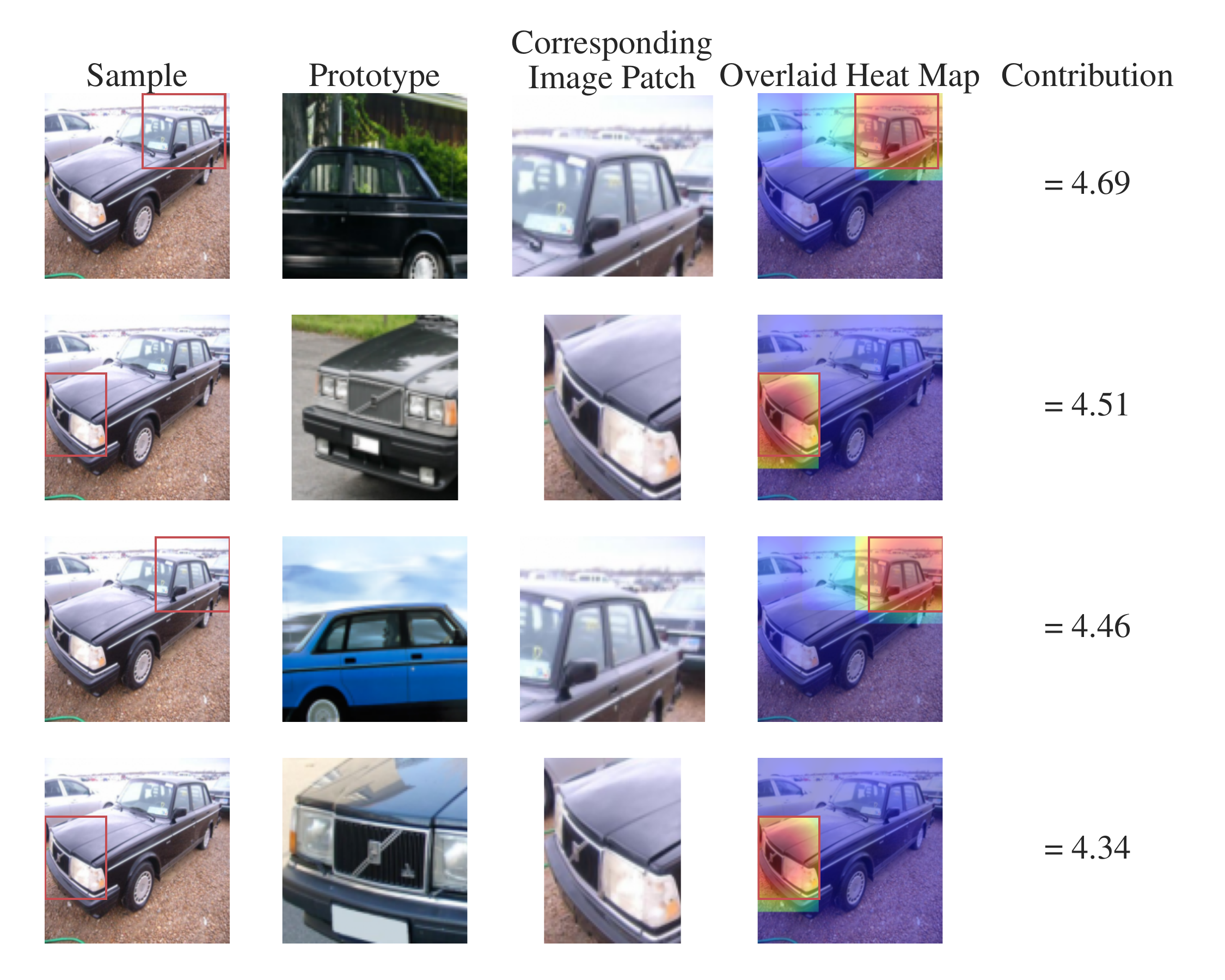}
    \end{subfigure}\\%
    \begin{subfigure}[b]{.4\linewidth}%
        \centering
        \includegraphics[width=\linewidth]{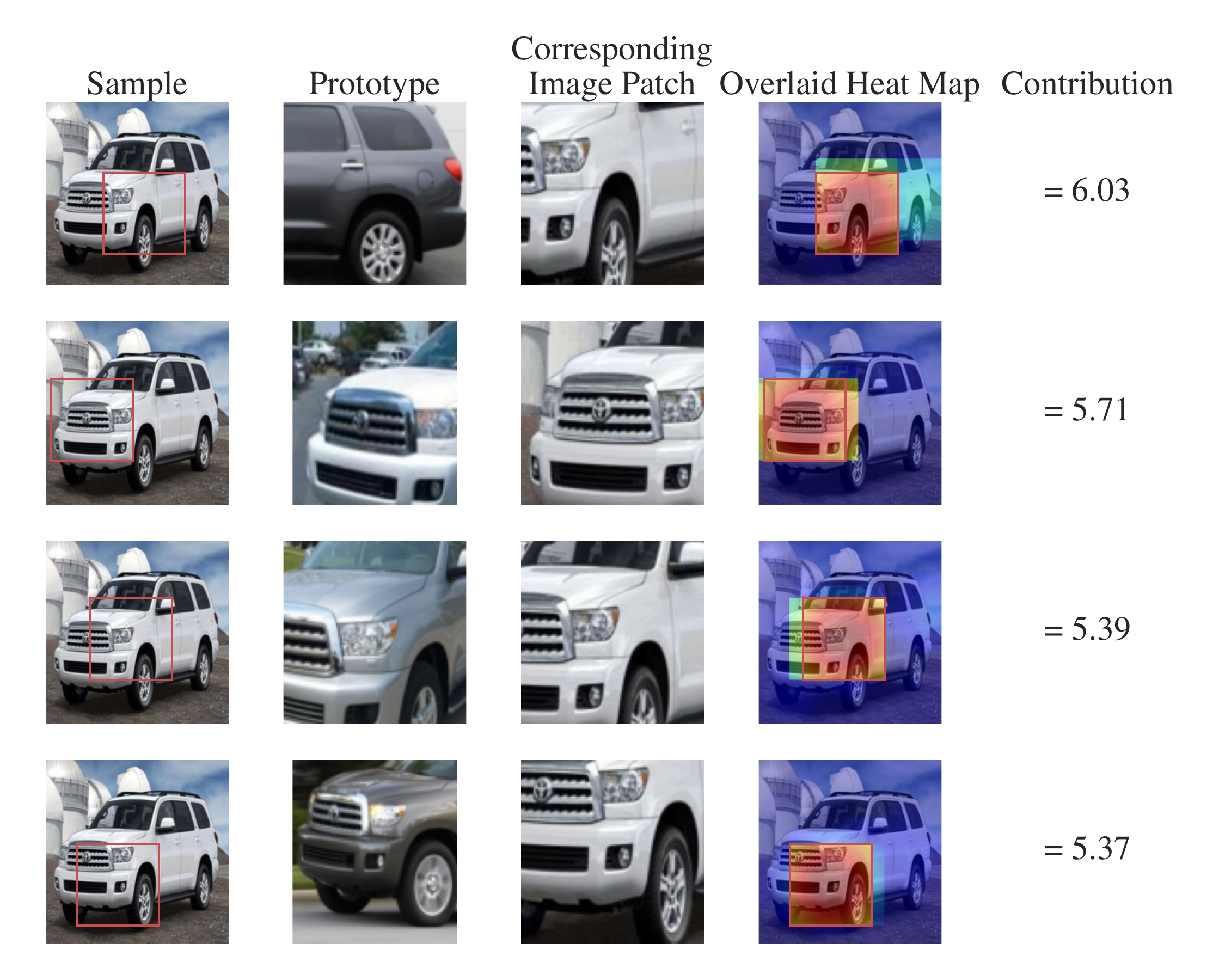}
    \end{subfigure}%
    \begin{subfigure}[b]{.4\linewidth}%
        \centering
        \includegraphics[width=\linewidth]{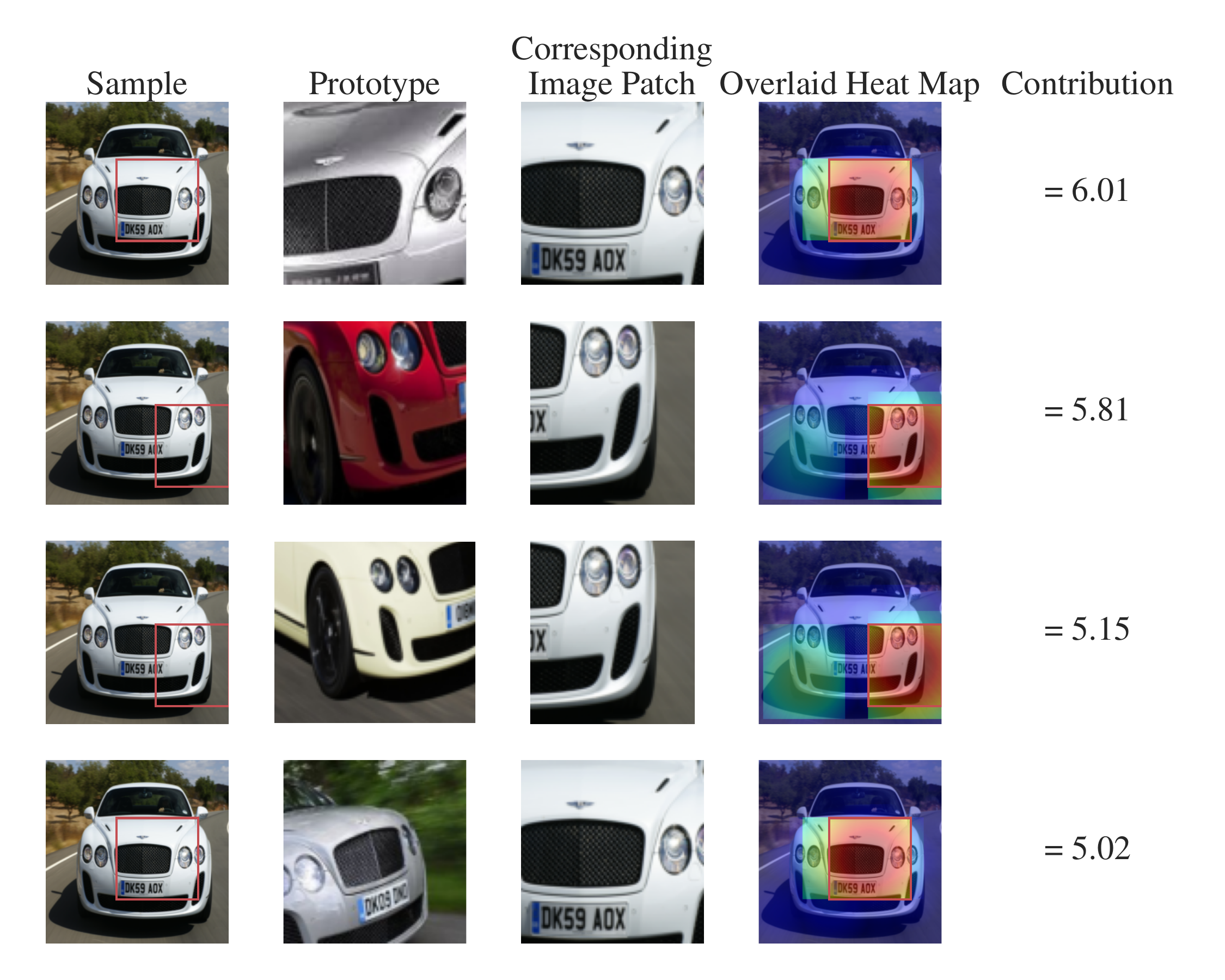}
    \end{subfigure}\\%
    \begin{subfigure}[b]{.4\linewidth}%
        \centering
        \includegraphics[width=\linewidth]{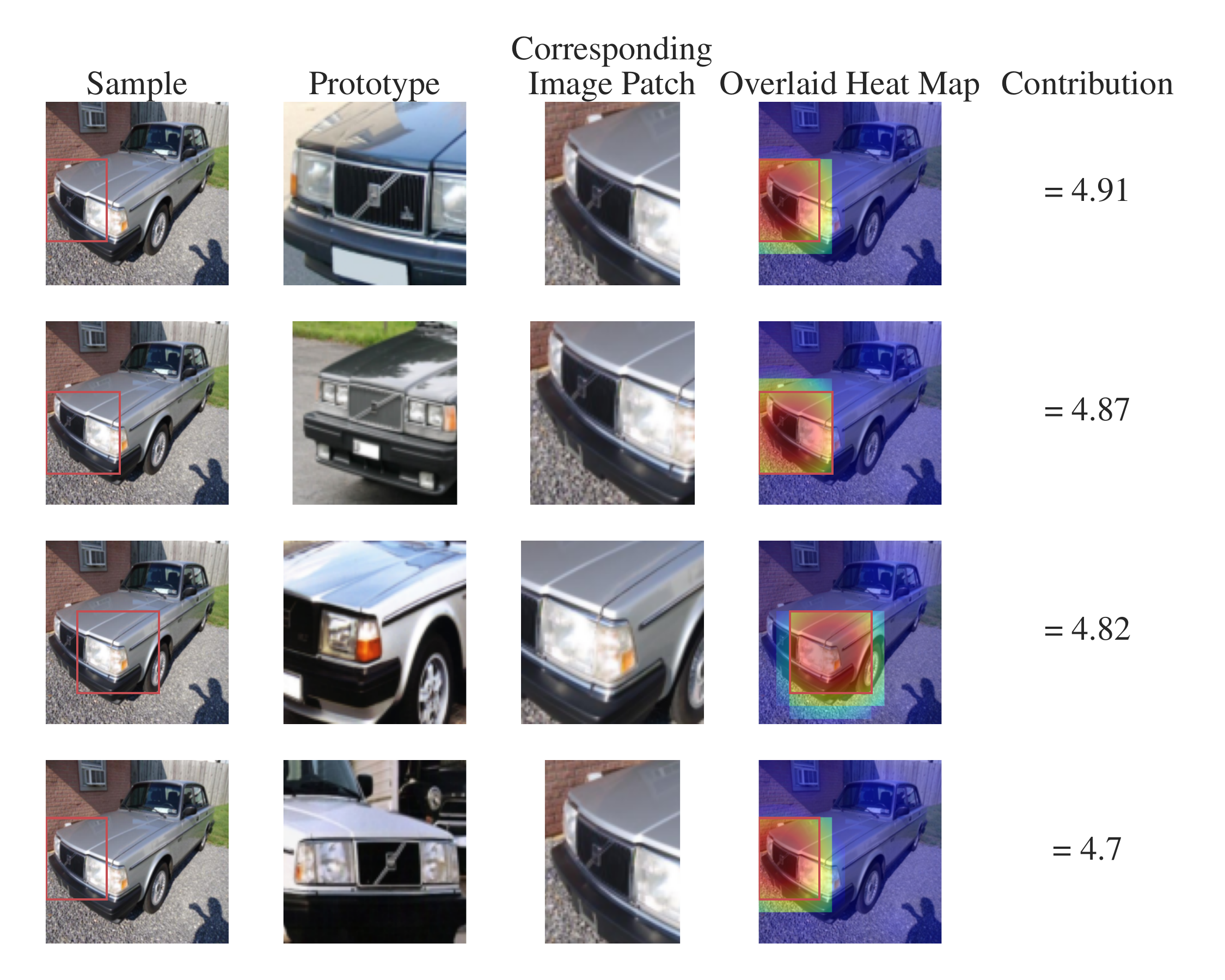}
    \end{subfigure}%
    \begin{subfigure}[b]{.4\linewidth}%
        \centering
        \includegraphics[width=\linewidth]{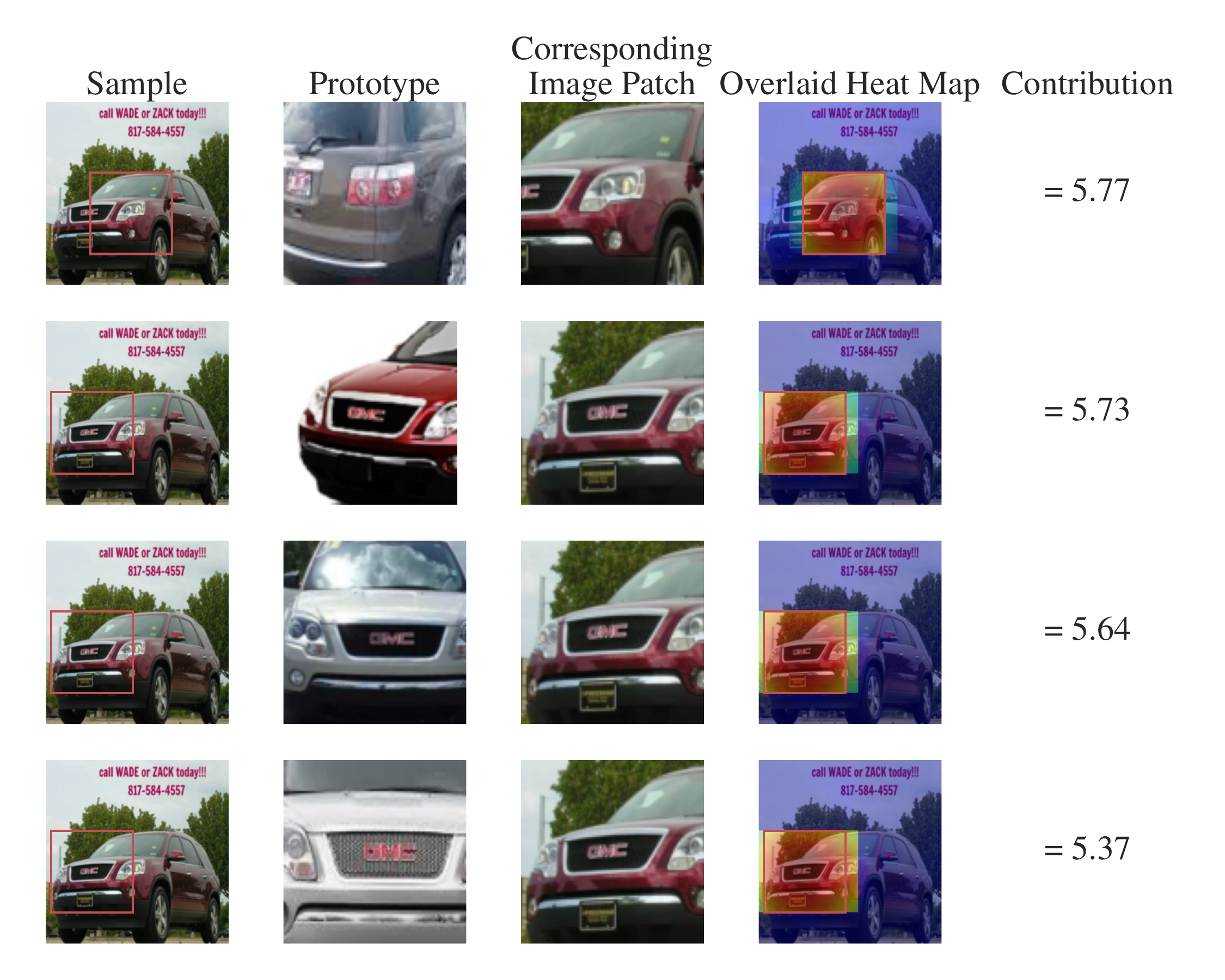}
    \end{subfigure}\\%
    \begin{subfigure}[b]{.4\linewidth}%
        \centering
        \includegraphics[width=\linewidth]{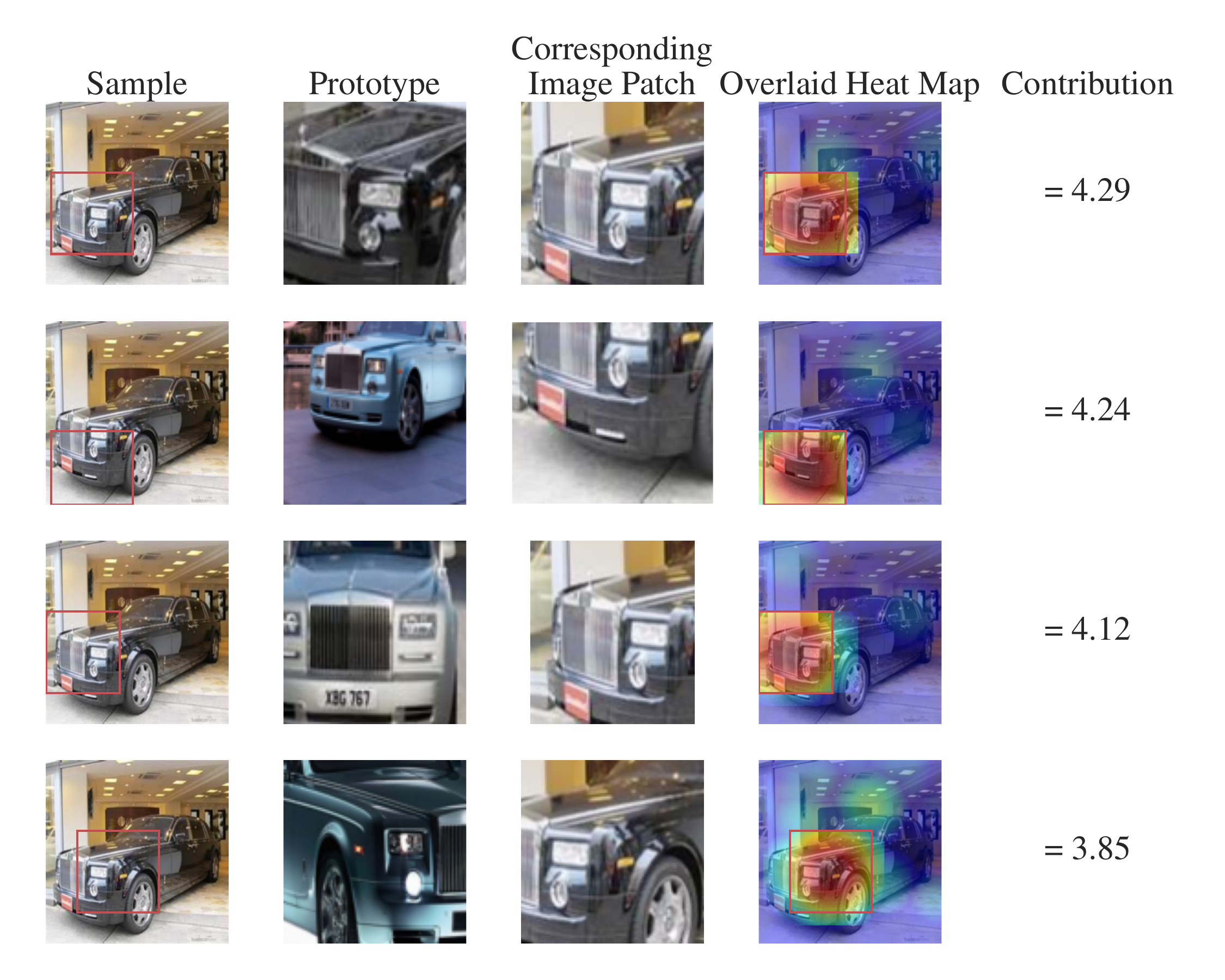}
    \end{subfigure}%
    \begin{subfigure}[b]{.4\linewidth}%
        \centering
        \includegraphics[width=\linewidth]{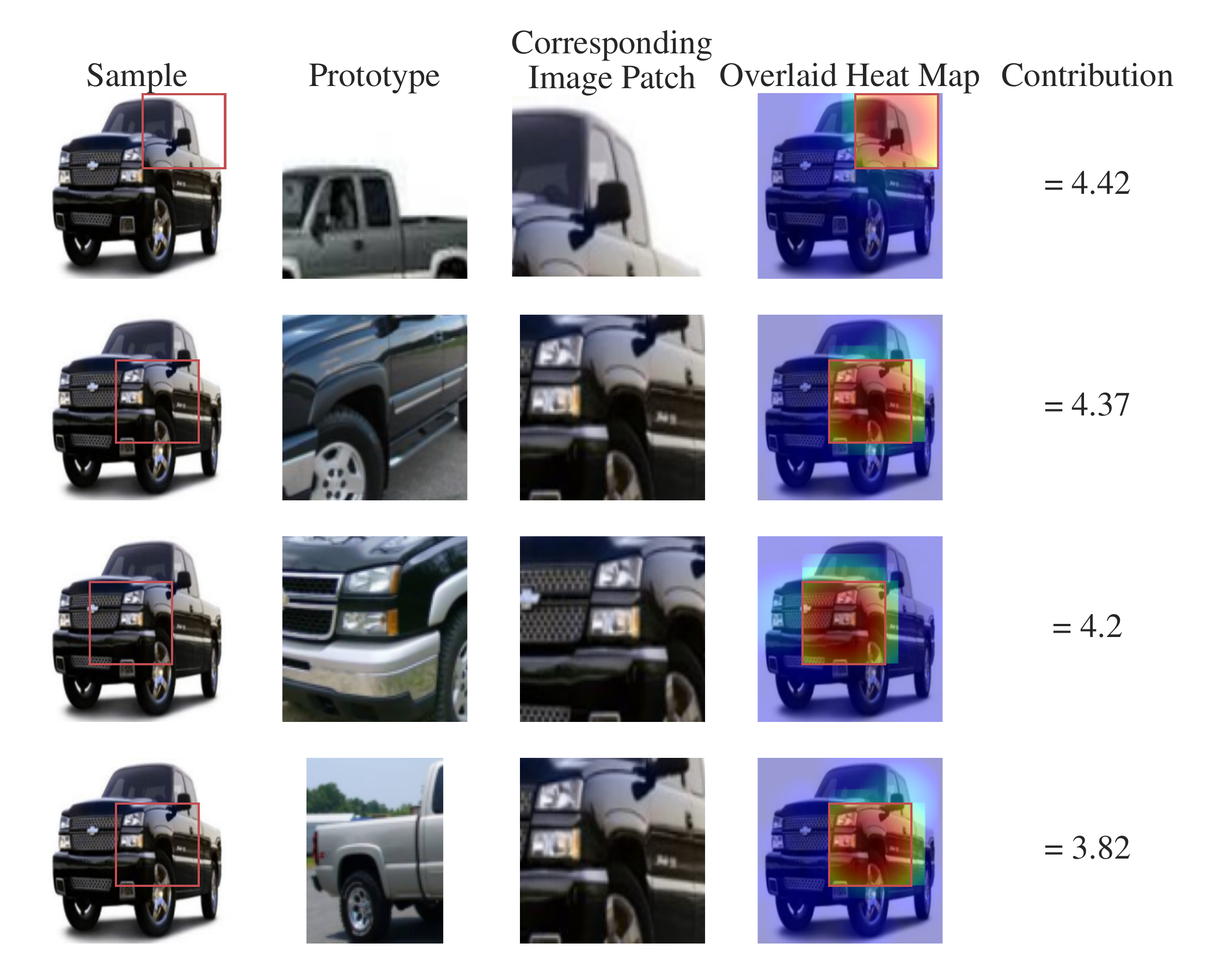}
    \end{subfigure}%
    \vspace{-2ex}
    \caption{Examples of \Ours{} explanations on Stanford Cars.}
    \label{fig:more-explanations-cars}
\end{figure*}

Figures~\ref{fig:pareto-cub} and \ref{fig:pareto-cars} present the discovered ``Goldilocks'' zone backbones evaluated in \Ours{} for CUB-200-2011 and Stanford Cars, respectively. As can be seen, the Pareto front is mostly retained, demonstrating that the ImageNette approach is a good proxy for backbone selection for \Ours{}. The results for various ImageNet-pre-trained \Protonets{} are provided in Table~\ref{tab:cars}.
Figures~\ref{fig:more-explanations} and \ref{fig:more-explanations-cars} shows more examples of explanations on CUB-200-2011 and Stanford Cars, respectively. We also include a video with the supplemental material submission that demonstrates the relevance ordering test for the CUB-200-2011 dataset.

\begin{figure}
    \centering
    \includegraphics[width=\linewidth]{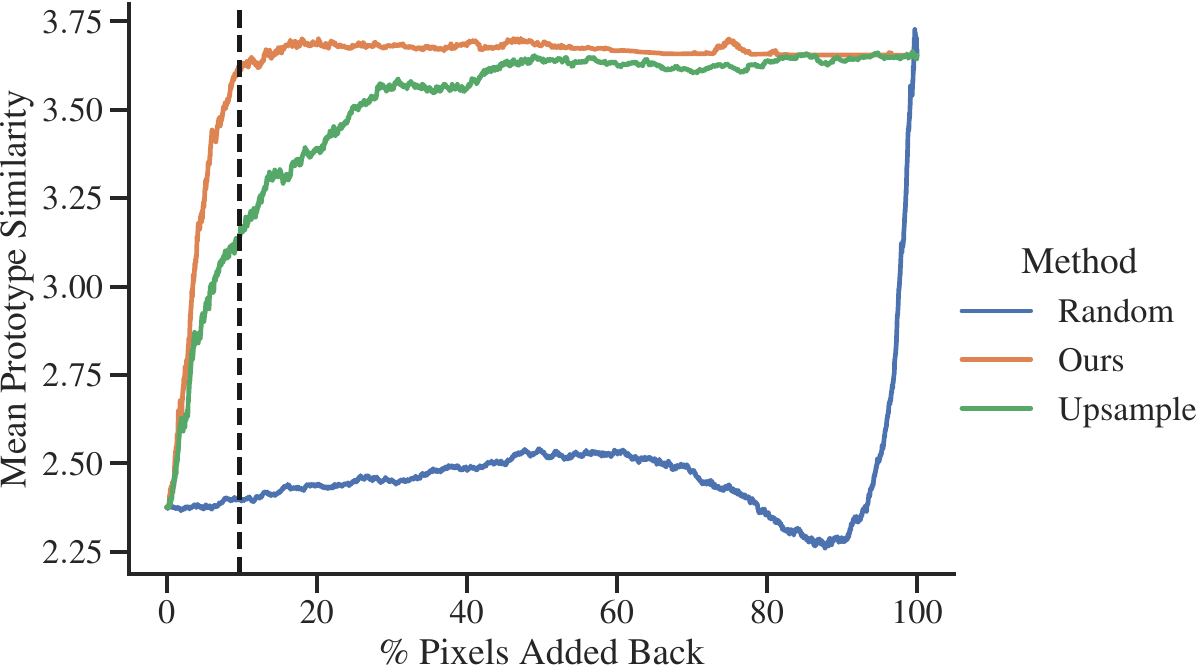}
    \caption{Relative ordering test for a VGG13@maxpool4. The original, random, and our pixel space mappings are compared. We achieve the highest AUSC and the lowest \%2R. \PRP{} is not shown because its implementation only supports ResNet models. The black dashed line indicates the mean receptive field of maxpool4.}
    \label{fig:rot-vgg13}
\end{figure}

\begin{figure}
    \centering
    \includegraphics[width=\linewidth]{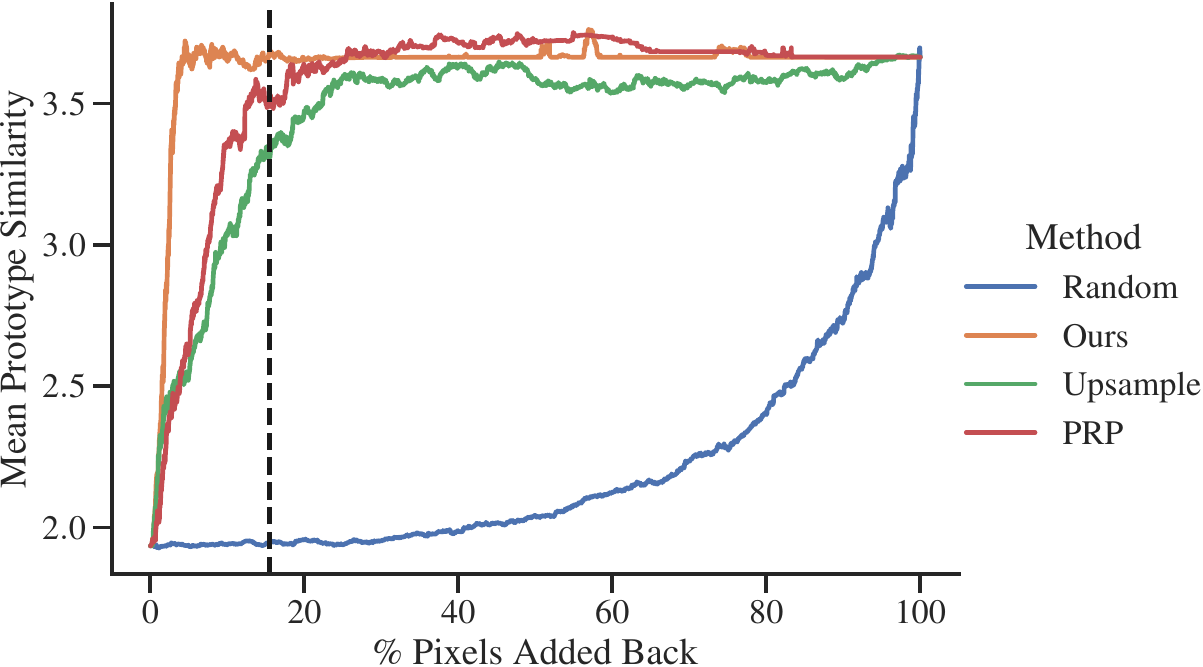}
    \caption{Relative ordering test for a ResNet18@layer2. The original, random, and our pixel space mappings are compared. We achieve the highest AUSC and the lowest \%2R. The black dashed line indicates the mean receptive field of layer2.}
    \label{fig:rot-resnet18}
\end{figure}

\begin{table*}
    \footnotesize
    \centering
    \renewcommand{\arraystretch}{1.2}
    \addtolength{\tabcolsep}{-1pt}
    \begin{tabular}{@{}ccccllrrrrrrcc@{}}
\toprule
       BBox & D1 & D2 & D3 & Model          &          $f$ & \makecell[r]{Expl.\\Size $+$} & \makecell[r]{Expl.\\Size $\pm$} &    $P$ & MRF & Accuracy &  $\pm$ & \makecell{Code\\Avail.} & \makecell{Val.\\Set} \\
\midrule
\multirow{7}{*}{\xmarkInv} & \multirow{3}{*}{\cmark} & \multirow{3}{*}{\cmark} & \multirow{3}{*}{\cmark}
& \TabOurs{} &    VGG19{\scriptsize @maxpool5} &                           \textbf{10} &                           \textbf{10} &  2000 &  70.4 &   \textbf{86.44} &     0.2 &                  \cmark &                    \cmark \\
& & & & \TabOurs{} &    VGG16{\scriptsize @maxpool5} &                           \textbf{10} &                           \textbf{10} &  2000 & 52.5 &    84.94 &     0.1 &                  \cmark &                    \cmark \\
& & & & \TabOurs{} &    VGG13{\scriptsize @maxpool5} &                           \textbf{10} &                           \textbf{10} &  2000 & 33.5 &    81.72 &     0.2 &                  \cmark &                    \cmark \\
& & & & \TabOurs{} &    VGG16{\scriptsize @maxpool4} &                           \textbf{10} &                           \textbf{10} &  2000 &  15.8 &   80.05 &     0.3 &                  \cmark &                    \cmark \\\cline{2-14}
 & \multirow{3}{*}{\cmark} & \multirow{3}{*}{\xmark} & \multirow{3}{*}{\cmark} &\TabSupportProtoPNet{} \cite{ST-ProtoPNet} &    ResNet152 &                           180 &                           35280 &  17640 &  100 &   \textbf{87.30} &     -- &                  \xmark &                    ? \\
       &        &        &        & \textsc{Deformable} \cite{DeformableProtoPNet} &    ResNet152 &                           180 &                           35280 &  17640 &  100 &   86.50 &     -- &                  \cmark &               \xmark \\
       &        &        &        & \TabSTProtoPNet{} \cite{ST-ProtoPNet} &    ResNet152 &                            20 &                            3920 &   1960 &  100 &   85.30 &     -- &                  \xmark &                    ? \\
\midrule
\multirow{7}{*}{\cmarkInv} & \multirow{7}{*}{\cmark} & \multirow{7}{*}{\xmark} & \multirow{7}{*}{\cmark} & \TabSTProtoPNet{} \cite{ST-ProtoPNet} &  DenseNet161 &                            20 &                            3920 &   1960 &   100 &  \underline{92.70} &    0.2 &                  \xmark &                    ? \\
       &        &        &        & \TabTesNet{} \cite{TesNet} &  DenseNet161 &                            20 &                            3920 &   1960 &   100 &  \underline{92.60} &    0.3 &                  \cmark &               \xmark \\
       &        &        &        & \TabProtoPool{} \cite{ProtoPool} &     ResNet34 &                            20 &                             390 &    195 &  100 &   89.30 &    0.1 &                  \cmark &               \xmark \\
       &        &        &        & \TabProtoPNet{} \cite{protopnet} &        VGG19 &                            20 &                            3920 &   1960 &   100 &  87.40 &    0.3 &                  \cmark &               \xmark \\
       &        &        &        & \TabProtoTree{} \cite{ProtoTree} &     ResNet34 &                            22 &                             390 &    195 &   100 &  86.60 &    0.2 &                  \cmark &               \xmark \\
       &        &        &        & \TabProtoPShare{} \cite{ProtoPShare} &     ResNet34 &                           960 &                             960 &    480 & 100 &    86.38 &     -- &                  \cmark &               \xmark \\
       &        &        &        & \TabProtoToProto{} \cite{Proto2Proto} &     ResNet18 &                            20 &                            3920 &   1960 &  100 &   84.00 &     -- &                  \cmark &               \xmark \\
\midrule
\multirow{2}{*}{\xmarkInv} & \xmark & \xmark & \xmark & \TabViTNeT{} \cite{ViT-NeT} &      SwinT-B &                            {12} &                             {126} &     63 &  100 &   \textbf{95.00} &     -- &                  \cmark &               \xmark \\\cline{2-14}
       &  \xmark & \xmark    & \cmark & \TabProtoPFormer{} \cite{ProtoPFormer} &  CaiT-XXS-24 &                            30 &                            5880 &   2940 &  100 &   91.04 &     -- &                  \cmark &               \xmark \\
\bottomrule
\end{tabular}
\addtolength{\tabcolsep}{1pt}
\caption{\Protonet{} results on Stanford Cars with ImageNet used for pre-training. Columns D1, D2, and D3 correspond to the three desiderata established in the main text.
``BBox'' indicates whether a method crops each image using a bounding box annotation.
The best results of \Protonets{} with and without such annotations are \textbf{bold} and \underline{underlined}, respectively.  Best results that are within one standard deviation of each other are all emphasized. The table is split based on whether the method meets at least two desiderata.}
\label{tab:cars}
\end{table*}

\section{Additional Receptive Field Algorithm Details}

\begin{figure}
    \centering
    \includegraphics[width=\linewidth]{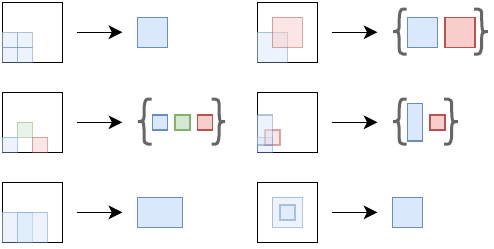}
    \caption{Examples of index merging with 2D slices. Slices that cannot be merged are retained as a set.}
    \label{fig:hc-merge}
\end{figure}

Here, we provide addition description of are function receptive field computation algorithm, \RFAlg{}.
Some complexity of the algorithm comes from handling arbitrary non-sequential architectures.
\RFAlg{}, takes a neural network as input and outputs the \textit{exact} receptive field of every neuron in the neural network.
Recall that a neuron is a function of a \textit{subset} of pixels defined by its receptive field.
\RFAlg{} represents receptive fields as hypercubes (multidimensional tensor slices).
For convolutional neural networks, we have four dimensions, but the batch size can safely be ignored.
We use the notation $\llbracket a,b \rrbracket$ to denote the slice (discrete interval) between $a$ and $b$.
The computation is outlined (see Algorithm~\ref{alg:rf-comp}) for image data for simplicity -- the algorithm works for any number of dimensions or type of data.
Given the directed acyclic computation graph of a neural network $\gG$, we can traverse the topologically sorted graph node by node to satisfy input dependencies. At the start, we initialize the receptive field
(\texttt{rf} attribute)
of the input node $v_0 \in \gG$ as $1 \times 1 \times 1$ slices into the input image.
Each consecutive node $v_k$ performs operation-dependent indexing into the receptive fields of its incoming nodes $\mathcal{U}_k$.
This indexing is operation-dependent and is encapsulated by the function \texttt{take\_from$(\cdot)$}.
For instance, the slices for a 2D convolution with a $5{\times} 5$ kernel, stride of 1, and $c_{\text{in}}$ channels at output position $3, 3$ would be $\{\{\llbracket 1,c_{\text{in}} \rrbracket, \llbracket 1,5 \rrbracket,\llbracket 1,5 \rrbracket\}\}$ where $\llbracket a,b \rrbracket$ denotes the slice between $a$ and $b$.
After,
we 
merge (\texttt{merge$(\cdot)$}) as many hypercubes as possible into a larger hypercube (consider, \eg{}, one hypercube inside of another)
to greatly reduce the space and time complexities. Figure~\ref{fig:hc-merge} gives several examples of this merge operation.
Finally, the receptive field-augmented graph $\gG$ is returned.

\begin{algorithm}[h]%
    \small
    \KwIn{$\gG$, the computation graph}
    \KwOut{$\gG$, augmented with receptive field information}
    Topologically sort $\gG$\;
    \tcp{Initialize each input node rf element as a $1{\times}1{\times}1$ hypercube}
    $v_0.\texttt{rf}_{dij} \gets \{\{\llbracket 1,3 \rrbracket, \llbracket i,i \rrbracket,\llbracket j,j \rrbracket\}\}$\;
    \For{$v_k \in \gG$}{
        $\mathcal{U}_k \gets $ all incoming nodes of $v_k$\;
        \tcp{Record slicing operations of $v_k$}
        $v_k.\texttt{rf}_{dij} \gets v_k.\texttt{take\_from}(\{u_k.\texttt{rf} \mid u_k \in \mathcal{U}_k\})$\;
        \tcp{Merge hypercubes}
        $v_k.\texttt{rf}_{dij} \gets \texttt{merge}(v_k.\texttt{rf}_{dij})$\;
    }
    \KwRet{$\gG$}
    \caption{{\RFAlg{}}($\gG$)}
    \label{alg:rf-comp}
\end{algorithm}%

\section{Additional Pixel Space Mapping Algorithm Details}

In order to compute a pixel space heat map, we propose an algorithm based on \RFAlg{} rather than naively upsampling an embedding space similarity map $\simmap_{ij}$.
Our approach uses the same idea as going from embedding space to pixel space.
Each pixel space heat map $\mM_{ij} \in \sR^{H\times W}$ is initialized to all zeros ($\mathbf{0}^{H \times W}$), and corresponds to a sample $\sample_i$ and a prototype $\prototype_j$.
Let $\mM_{ij}^\emS$ be the region of $\mM_{ij}$ defined by the receptive field of similarity score $\emS \in \simmap_{ij}$.
For each $\emS$, the pixel space heat map is updated as $\mM_{ij}^\emS \gets \max(\mM_{ij}^\emS, \emS)$ where $\max(\cdot)$ is an element-wise maximum that appropriately handles the case of overlapping receptive fields.
Note that we weight $\emS$ by $T(|r_i|, |r_j|, \sigma_{T})$ (via broadcasting) where $T$ generates a 2D Gaussian kernel, $|\cdot|$ gives the length of a discrete interval, and $\sigma_{T}$ is the standard deviation of the kernel. The first two arguments denote the height and width of the kernel, respectively. We set $\sigma_{T}$ to the larger of the height and width. The intuition behind this approach is that a receptive field is actually Gaussian~\cite{LuoERF2016} -- pixels in the center are more important, and pixels at the periphery are less important. We stress that this does not affect localization and the pixel space heat maps are largely unchanged without this step. However, this does affect the relevance order testing -- without this Gaussian weighting step, the pixels within a receptive field would all have the same value, so the most important pixel in the region is selected arbitrarily. Selecting in the center and ``spiraling'' outwards is more faithful to what is known about deep neural networks~\cite{LuoERF2016}. In experiments, we also confirm that this step improves the relevance ordering test scores, especially in the case of large receptive fields.
The full procedure ({\small\texttt{RFPixelSpaceMapping}}) is shown in Algorithm~\ref{alg:rf-heatmap}.

\begin{algorithm}[h]%
    \small
    \KwIn{$v_k$, the embedding layer in $\gG$ (augmented by Algorithm~\ref{alg:rf-comp})}
    \KwIn{$\simmap$, the similarity map for a prototype}
    \KwOut{$\mM$, a pixel space heat map}
    $\mM \gets $ matrix of zeros $\in \sR^{H \times W}$\;
    \For{\texttt{\textup{rf}}$_{dij} \in v_k.\texttt{\textup{rf}}$}{
        \For{$(r_d, r_i, r_j) \in \texttt{\textup{rf}}_{dij}$}{
            $\mM_{r_d r_i r_j} \gets \textup{max}(\simmap_{dij} \times T(|r_i|, |r_j|, \sigma_{T}), \mM_{r_d r_i r_j})$\;
        }
    }
    \KwRet{$\mM$}
    \caption{{\small\texttt{RFPixelSpaceMapping}}($v_k$, $\simmap$)}
    \label{alg:rf-heatmap}
\end{algorithm}%

\section{Experiment Setup and Reproducibility}

\paragraph{Hardware and Software}
The code for this paper was implemented in Python and primarily relies on PyTorch~\cite{paszke2019pytorch} and PyTorch Lightning~\cite{Falcon_PyTorch_Lightning_2019}. The original \ProtoPNet{}~\cite{protopnet} code was used to guide some development, but the majority of code deviated (especially the training code which actually uses a validation split for validation and tuning).
All experiments were run on NVIDIA A10 GPUs.
Unless stated elsewhere, all experimental results are reported as the average across 10 trials.
Our code will be made fully available upon publication.

\paragraph{Data Augmentation}

For the CUB-200-2011 and Stanford Cars datasets, we perform the following augmentations to each training image:
\begin{enumerate}
\item Resize smallest dimension of image to 510 pixels with bilinear interpolation
\item Gaussian blur with 5 $\times$ 5 kernel and random sigma in the range $[0.1,5]$ (1/10 probability)
\item Randomly adjust sharpness by a factor of 1.5 (1/10 probability)
\item Randomly rotate image in the range $[-15,15]$ degrees (1/3 probability)
\item Randomly distort the image perspective with a scale of 0.2 (1/3 probability)
\item Randomly shear the image by 10 degrees (1/3 probability)
\item Randomly flip the image horizontally (1/2 probability)
\item Randomly crop the image to 384 $\times$ 384 pixels
\item Downsample the image to 224 $\times$ 224 pixels with bilinear interpolation
\item Normalize the image to have the channel-wise mean and standard deviation of the full data set
\end{enumerate}
\noindent
For ImageNette, steps 1 and 9 are omitted, and the random crop is done to 224 $\times$ 224 pixels directly.

We use an augmentation factor which indicates how many augmented samples are generated for each training image. In \ProtoPNet{}, 30 augmentations are generated for each sample. In addition, \ProtoPNet{} uses offline (static) augmentation, whereas we use online augmentation.

\paragraph{Training}
The training procedure largely follows that of \ProtoPNet{}. For a warm-up period, we train just the add-on layers $\addonf$ and the prototypes $\prototypes$. Thereafter, all layers are trained. We use an exponential warm-up of the learning rate and use a cosine annealing learning rate scheduler (without restarts)~\cite{loshchilov2016sgdr}. Every $k$ epochs, we perform the prototype replacement procedure.

\paragraph{Hyperparameters} All hyperparameters are listed in the proceeding table.

\begin{table}
    \footnotesize
    \centering
    \begin{tabular}{@{}rl@{}}
        \toprule
        \textbf{Name} & \textbf{Value} \\
        \midrule
        Augmentation Factor & 16 \\
        Validation Set Proportion & 0.1 \\
        Pre-training & ImageNet \\
        $\lossClsCoef$ & 0 \\
        $\lossSepCoef$ & 0 \\
        $\distancef$ & Cosine Distance \\
        $\varepsilon$ & $10^{-6}$ \\
        $P$ & $C \times 10$ \\
        $D$ & 192 \\
        $H_{p}$ & 1 \\
        $W_{p}$ & 1 \\
        Learning Rate ($\backbone$ parameters) & 0.0001 \\
        Learning Rate ($\addonf$ parameters) & 0.003 \\
        Learning Rate ($\prototypes$) & 0.003 \\
        Warm-Up Epochs & 5 \\
        Learning Rate Scheduler (Warm-Up) & Exponential \\
        Learning Rate Scheduler & Cosine \\
        Weight Decay (All Parameters Except $\prototypes$) & 0.001 \\
        Optimizer & Adam \\
        Batch Size & 64 \\
        Epochs & 20 \\
        Prototype Replacement & Every 4 Epochs \\
        \bottomrule
    \end{tabular}
    \caption{Table of \Ours{} hyperparameters used in experiments.}
    \label{tab:hparams}
\end{table}

\section{``Goldilocks'' Zone Experimental Details}

For this experiment, we follow the same training procedure as for \Ours{}, except for any \Protonet{}-specific training (\eg{}, prototype replacement). See the previous section for data augmentation details. We select pre-packaged and ImageNet pre-trained architectures from PyTorch~\cite{paszke2019pytorch} to evaluate the intermediate layers of. Recall that the goal of this experiment is to discover backbones suitable for \Ours{} by observing the Pareto front of the mean receptive field and accuracy on ImageNette~\cite{imagenette}. See the main text for discussion about this data set and justification for this approach. For each selected intermediate layer, the network is dissected at that point and a new classification head is appended which comprises a $1 \times 1$ 2D adaptive average pooling layer, a flattening of the unary dimensions, and a fully-connected layer. The training procedure is carried out for 10 epochs with a batch size of 32.
The proceeding table outlines the selected architectures and intermediate layers that are evaluated.

\begin{table}
    \footnotesize
    \centering
    \begin{tabular}{@{}lr@{}}
        \toprule
        \textbf{Architecture} & \textbf{Intermediate Layers} \\
        \midrule
         densenet121 & \multirow{4}{*}{\makecell[l]{conv0,norm0,relu0,pool0,denseblock1,\\transition1,denseblock2,transition2,\\denseblock3,transition3,denseblock4,norm5,\\avgpool,classifier}} \\
         densenet161 & \\
         densenet169 & \\
         densenet201 & \\[1ex]
         inception\_v3 & \makecell[l]{Conv2d\_1a\_3x3,\\Conv2d\_2a\_3x3,Conv2d\_2b\_3x3,maxpool1,\\Conv2d\_3b\_1x1,Conv2d\_4a\_3x3,maxpool2,\\Mixed\_5b,Mixed\_5c,Mixed\_5d,Mixed\_6a,\\Mixed\_6b,Mixed\_6c,Mixed\_6d,Mixed\_6e,\\Mixed\_7c,avgpool,fc} \\[1ex]
         resnet18 & \multirow{10}{*}{\makecell[l]{conv1,maxpool,layer1,layer2,layer3,layer4,\\avgpool,fc}} \\
         resnet34 & \\
         resnet50 & \\
         resnet101 & \\
         resnet152 & \\
         resnext101\_32x8d & \\
         resnext101\_64x4d & \\
         resnext50\_32x4d & \\
         wide\_resnet50\_2 & \\
         wide\_resnet101\_2 & \\[1ex]
         squeezenet1\_0 & \multirow{2}{*}{\makecell[l]{conv1,maxpool1,maxpool2,maxpool3,\\features,final\_conv}} \\
         squeezenet1\_1 & \\[1ex]
         vgg11 & \multirow{4}{*}{\makecell[l]{conv1,maxpool1,maxpool2,maxpool3,\\maxpool4,maxpool5,avgpool,classifier}} \\
         vgg13 & \\
         vgg16 & \\
         vgg19 & \\
        \bottomrule
    \end{tabular}
    \caption{All evaluated intermediate layers of backbone candidate architectures.}
    \label{tab:hackjobs}
\end{table}

\section{Similarity Function Formulation}

We reduce numerical error of the original similarity function by reformulating it as:
\begin{align}\label{eq:sim}
\simf(d) =
\underbrace{
    \log\left(
        \frac
        {d + 1}
        {d + \varepsilon}
    \right)
}_{\text{Original formulation}}
=
\underbrace{
    \log\left(
        \frac
        {1}
        {d + \varepsilon} + 1
    \right)
}_{\text{New formulation}}
.
\end{align}
To validate its accuracy, we compare its scores across various values to the expected similarity scores with infinite precision. We use the \texttt{mpmath}~\cite{mpmath} library to implement infinite precision. The table below compares the mean-squared-error of our approach to the original and our version of the similarity function. Our reformulation achieves lower error, especially in the $[1,10]$ value range for both 32- and 64-bit IEEE 754 floating point numbers.

\begin{table}[H]
    \footnotesize
    \centering
    \begin{tabular}{lrrrr}
    \toprule
        \multirow{2}{*}{\textbf{Dtype}} & \multirow{2}{*}{\textbf{Region}} & \multicolumn{2}{c}{\textbf{MSE}} & \multirow{2}{*}{\textbf{\% Improved}} \\
        &&\textbf{Original} & \textbf{Ours} & \\\midrule
        \multirow{5}{*}{Float32} & 0, 1e-6 & 1.05e-13 & 1.04e-13 & 1.11\% \\ 
         & 1e-6, 1e-3 & 4.08e-14 & 3.97e-14 & 2.69\% \\ 
         & 1e-3, 1 & 3.56e-15 & 3.30e-15 & 7.41\% \\ 
         & 1, 10 & 1.42e-15 & 1.03e-15 & 27.13\% \\ 
         & 10, 1000 & 1.19e-15 & 1.18e-15 & 0.90\% \\
        \multirow{5}{*}{Float64} & 0, 1e-6 & 3.61e-31 & 3.56e-31 & 1.14\% \\ 
         & 1e-6, 1e-3 & 1.40e-31 & 1.35e-31 & 3.09\% \\ 
         & 1e-3, 1 & 1.16e-32 & 1.05e-32 & 9.12\% \\ 
         & 1, 10 & 4.63e-33 & 3.27e-33 & 29.28\% \\ 
         & 10, 1000 & 4.12e-33 & 4.09e-33 & 0.65\% \\ 
         \bottomrule
    \end{tabular}
    \caption{The comparative numerical error of the similarity functions: original and ours. Mean-squared-error (MSE) is shown for IEEE 754 floating point data types. Our reformulation is more accurate especially in the $[1,10]$ range of values.}
\end{table}

\section{Issues with \ProtoPNet{} Code Base}

Upon inspection of the original code base\footnote{\url{https://github.com/cfchen-duke/ProtoPNet}}, we discovered that the test set accuracy is used to influence training of \ProtoPNet{}.
In fact, neither \ProtoPNet{} nor its extensions for image classification that are mentioned in the paper employ a validation set in provided implementations.
Part of this is due to some code bases being derived from the original implementation.
\ProtoPNet{} peeked at test set, which propagated to subsequent paper implementations. According to their implementation, they did not only have no validation set,
In addition, the provided code also used the accuracy on the test set to influence when training should stop.

The relevant portions of code are shown below\footnote{These verbatim snippets are taken from the git commit \texttt{c02e8568900f20df704f65aeb86f0dd1738ca785} (most recent commit as of 2023-03-13)}. In \texttt{main.py}, there is a training loop that saves the model twice each epoch (once before prototype replacement and once after). Each time that the model is saved, the accuracy on the test set is stored with the model. In addition, if the test accuracy is above 70\%, then a message is logged to the console stating that the test accuracy is above said target.

{\scriptsize
\begin{lstlisting}[language=Python,numbers=left,firstnumber=172,breaklines=true,caption=\texttt{main.py} Snippet]
accu = tnt.test(model=ppnet_multi, dataloader=test_loader,
                class_specific=class_specific, log=log)
save.save_model_w_condition(model=ppnet, model_dir=model_dir, model_name=str(epoch) + 'push', accu=accu,
                            target_accu=0.70, log=log)
\end{lstlisting}
}

The \texttt{save} module is from the \texttt{save.py}. The relevant snippet is shown below.

{\scriptsize
\begin{lstlisting}[language=Python,numbers=left,firstnumber=4,breaklines=true,caption=\texttt{save.py} Snippet]
def save_model_w_condition(model, model_dir, model_name, accu, target_accu, log=print):
    '''
    model: this is not the multigpu model
    '''
    if accu > target_accu:
        log('\tabove {0:.2f}%'.format(target_accu * 100))
        # torch.save(obj=model.state_dict(), f=os.path.join(model_dir, (model_name + '{0:.4f}.pth').format(accu)))
        torch.save(obj=model, f=os.path.join(model_dir, (model_name + '{0:.4f}.pth').format(accu)))
\end{lstlisting}
}

The purpose of the held-out test set is to properly measure generalization error, not be used to tune hyperparameters nor influence training. Doing so is actually overfitting the distribution of the test set rather than truly improving performance. This phenomenon unfortunately has been long-standing in deep learning research, affecting progress with prominent datasets, including CIFAR-10 and ImageNet~\cite{recht2019imagenet,recht2018cifar,hendrycks2021natural}. In our implementation, we employ a proper validation set and tune hyperparameters only according to accuracy on this split. This ensures that we are properly measuring generalization error.

In addition, we were able to approach but not reproduce the originally reported accuracies, nor could some others, using the provided code and instructions\footnote{See \texttt{ProtoPNet} GitHub issue numbers 9, 10, 11 (the authors did not respond) and \cite{ExplainingPrototypes}.}.

\section{Full Discussion of \ProtoPNet{} Variants and Extensions}

The idea of sharing prototypes between classes has been explored in
\ProtoPShare{}~\cite{ProtoPShare} (prototype merge-pruning) and
\ProtoPool{}~\cite{ProtoPool} (differential prototype assignment).
In \ProtoTree{}~\cite{ProtoTree}, the classification head is replaced by a differentiable tree, also with shared prototypes.
A procedure is also proposed to convert to a hard tree to improve interpretability.
An alternative embedding space is explored in \TesNet{}~\cite{TesNet} based on Grassmann manifolds.
The authors additionally propose a new similarity function, an orthogonality loss, and a subspace loss.
A \Protonet{}-specific knowledge distillation approach is proposed in \ProtoToProto{}~\cite{Proto2Proto} by enforcing that student prototypes and embeddings should be close to those of the teacher.
\DeformableProtoPNet{}~\cite{DeformableProtoPNet} extends the \ProtoPNet{} architecture with deformable prototypes.
\STProtoPNet{}~\cite{ST-ProtoPNet} learns support prototypes that lie near the classification boundary and trivial prototypes that are far from the classification boundary.

\begin{figure}[H]
    \centering
    \includegraphics[width=\linewidth]{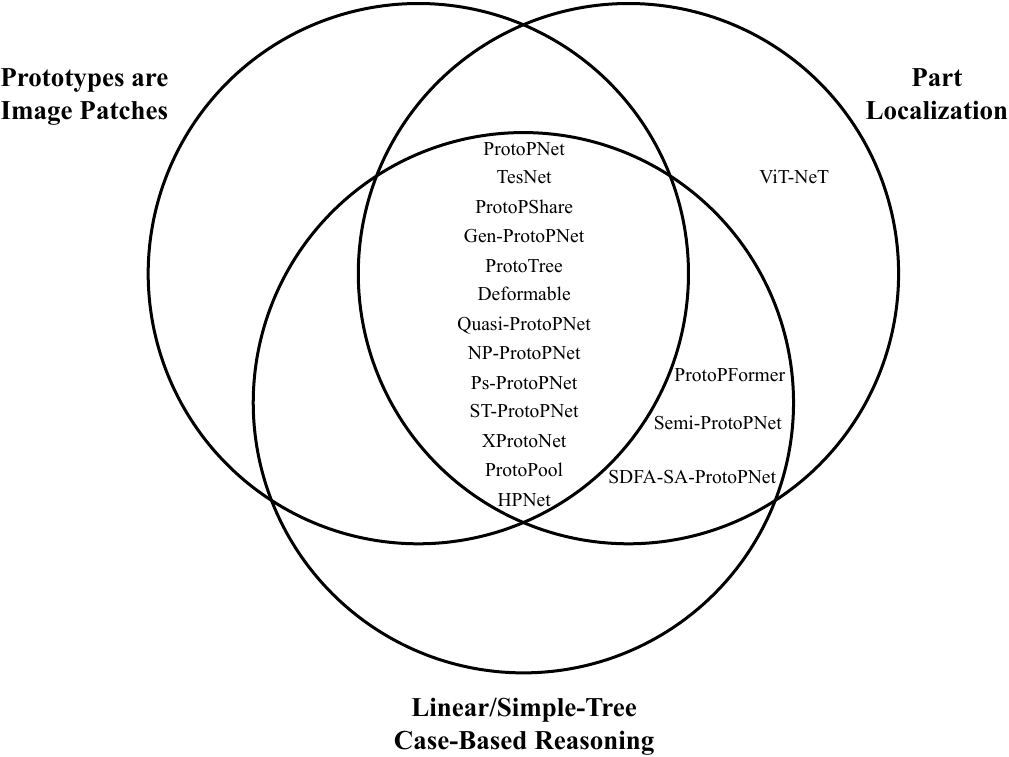}
    \caption{Proposed desiderata of \Protonets{}: part localization, linear/simple-tree case-based reasoning, and prototypes are image patches. Architectures at the intersection of all three desiderata are considered \textit{true \Protonets{}}. Only architectures for image classification tasks are shown.}
    \label{fig:ppnet-venn}
\end{figure}

\ProtoPNet{} has also been extended for graphs.
\ProtGNN{}~\cite{ProtGNN} is an adaptation for graph classification and is built on by \PxGNN{}~\cite{PxGNN}, which extends it for node classification.
Simultaneously, \GCNTesNetProtoPNet{}~\cite{GCNTesNetProtoPNet} was also proposed for graph and node classification.
Likewise \ProSeNet{}~\cite{ProSeNet}, extends the architecture for sequential data
and \ProtoSeg{}~\cite{ProtoSeg}
can handle supervised image segmentation tasks.
\ProtoPNet{} has been adapted to task-specific applications, including EEG data (\ProtoPMedEEG{}~\cite{ProtoPMed-EEG}), Earth science data (\ProtoLNet{}~\cite{ProtoLNet}), 
and deepfake detection (\DPNet{}~\cite{DPNet}).
There have been a handful of \ProtoPNet{} extensions tailored for diagnosing the chest CT scans of COVID-19 patients This includes
\QuasiProtoPNet{}~\cite{Quasi-ProtoPNet} (removes non-class weight connections in $h$),
\NPProtoPNet{}~\cite{NP-ProtoPNet} (fixes weights in $h$ to $+1$ and $-1$ for same- and non-class connections),
\GenProtoPNet{}~\cite{Gen-ProtoPNet} (uses larger prototype kernel sizes),
\PsProtoPNet{}~\cite{Ps-ProtoPNet} (combines \GenProtoPNet{} and \NPProtoPNet{}),
and
\XProtoNet{}~\cite{XProtoNet} (prototypes are compared with dynamically-sized feature patches).

\fakeparagraph{\Protonet{} Tools}
A few tools have also been proposed. In \ProtoPDebug{}~\cite{ProtoPDebug}, a concept-level debugger for \ProtoPNet{} is proposed in which a human supervisor, guided by model explanations, removes part-prototypes that have learned shortcuts or confounds.
In
\cite{ExplainingPrototypes}, a methodology is developed to enhance \Protonet{} explanations with color, hue, saturation, shape, texture, and contrast information.

In an attempt to improve \ProtoPNet{} visualizations, an extension of layer-wise relevance propagation~\cite{bach2015pixel}, Prototypical Relevance Propagation (\PRP{}), is proposed to create more model-aware explanations~\cite{PRP}. \PRP{} is quantitatively more effective in debugging erroneous prototypes and assigning pixel relevance than the original approach.

\paragraph{\Protonet{}-Like Methods}
The following papers are inspired by \ProtoPNet{} but cannot be considered to be the same class of model. This is due to not fulfilling at least one of the proposed \Protonet{} desiderata.

\ViTNeT{}~\cite{ViT-NeT} combines a vision transformer~(ViT) with a neural tree decoder that learns prototypes. However, its training
does not employ prototype pushing and has additional layers after the embedding layer that modify the embedding space. This breaks the mapping back to pixel space.

In another transformer-based approach, \ProtoPFormer{}~\cite{ProtoPFormer} exploits the inherent architectural features (local and global branches) of ViTs. The prototype layer has both local and global prototypes, and a focal loss concentrates local prototypes on heterogeneous regions of the foreground.
However, the training procedure of the method removes prototype replacement.

\SemiProtoPNet{}~\cite{Semi-ProtoPNet} fixes the readout weights as \NPProtoPNet{} does and is applied to classification of defective structures in power distribution networks. However, the training procedure of the method also skips over prototype replacement.

In \SDFASAProtoPNet{}~\cite{SDFA-SA-ProtoPNet}, a shallow-deep feature alignment (SDFA) module aligns the similarity structures between deep and shallow layers.
In addition, a score aggregation (SA) module aggregates similarity scores of prototypes in a class-wise manner to avoid learning inter-class information.
Notably, the authors attempt to quantitatively evaluate the interpretability of prototype-based explanations rather than relying on qualitative examples as many other extensions have done. We discuss the proposed metrics in the main text. Once again, the training procedure omits prototype replacement.

Throughout each of these works, the main justification for removing prototype replacement is that it harms task accuracy.

\ProtoVAE{}~\cite{ProtoVAE} is an extension of \ProtoPNet{} for variational auto-encoders (VAEs). It outperforms \ProtoPNet{} on a variety of classification tasks. It uses an orthogonality loss for intra-class diversity (same as \TesNet{}). However, it does not employ prototype replacement, opting to rather use a decoder used to visualize prototypes.

\paragraph{Retrospective on Extensions}
Despite all these efforts, all extensions still have fundamental issues with object part localization, pixel space grounding, and heat map visualizations. We discuss what these issues are in detail in the main text.

\section{Additional Consistency and Stability Details}

For the CUB-200-2011 dataset, there are 15 object parts: back, beak, belly, breast, crown, forehead, left eye, left leg, left wing, nape, right eye, right leg, right wing, tail, and throat. However, we treat the left and right versions of an object as the same object as 1) semantically, they are the same object, 2) we do not want to penalize \Protonets{} for learning invariance to flips or rotations, and 3) data augmentation flips images, which makes lefts look like rights (and vice versa).
In addition, owing to data preprocessing, some object parts are not visible in images that are visible in the uncropped image. Naturally, we do not penalize \Protonets{} for not matching with parts that are not visible.

\paragraph{Limitations}
The consistency and stability metrics ignore the localization capability of \Protonets{} and arbitrarily set the localization window to 72 $\times$ 72 pixels, centered around the localization bounding box midpoint. In addition, the threshold of 0.8 is arbitrary for a prototype to be consistent. We argue that a soft score (the average of the prototype-object coinciding frequencies without thresholding) makes far more sense. It can be interpreted as the average consistency of all prototypes, rather than the proportion of prototypes that are at least 80\% consistent. When computing these soft scores, we noticed that the majority of them were in the $[0.6,0.8]$ range. Last, the metric relies on human object part annotations -- in the case of CUB-200-2011, all annotations are of bird parts. However, it is well-known that prototypes learn backgrounds and other foreground objects that are also discriminatory (such as oceans, tree branches, and human hands)~\cite{protoStudy1,protoStudyHIVE}. Prototypes may be consistent (and stable), but associated with the wrong part of the image just because of the 1) limited part annotations, and 2) the arbitrary 72 $\times$ 72 pixel window.
Nonetheless, the metrics do allow for comparative interpretability evaluation between \Protonets{}.
Further information on the consistency and stability metrics is available in our code implementation as well as~\cite{SDFA-SA-ProtoPNet}.

\section{Additional Small Improvements}
We noticed that the magnitude of feature maps always dictates prototype similarity when using Euclidean distance. Using cosine distance eliminates the influence of magnitude on prototype similarity.

In \ProtoPNet{}, the incorrect gain and initialization method is used in the initialization of the convolutional weights in $\addonf$ before the sigmoid activation. The original implementation uses a Kaiming normal initialization with a gain that is intended for the ReLU activation. Instead, we use the Xavier normal initialization with a gain of one.